\documentclass[lettersize,journal]{IEEEtran}
\usepackage{algorithmic}
\usepackage{algorithm}
\usepackage{multirow}
\usepackage{amsmath,amssymb,amsfonts}
\usepackage{booktabs}
\usepackage{array}
\usepackage[caption=false,font=normalsize,labelfont=sf,textfont=sf]{subfig}
\usepackage{textcomp}
\usepackage{stfloats}
\usepackage{url}
\usepackage{verbatim}
\usepackage{graphicx}
\usepackage{cite}
\usepackage{subcaption}
\usepackage{booktabs}
\usepackage{enumitem}
\usepackage[table]{xcolor}
\usepackage{tabularx}
\usepackage[most]{tcolorbox}
\usepackage[dvipsnames]{xcolor}

\newtcolorbox{findingbox}[1][]{%
    colframe=black,
    colback=white,
    boxrule=0.5pt,
    arc=2pt,
    left=5pt,
    right=5pt,
    top=3pt,
    bottom=3pt,
    boxsep=0pt,
    before skip=8pt,
    after skip=8pt,
    fontupper=\itshape,
    coltitle=black,
    title={#1},
    attach boxed title to top left={xshift=5pt,yshift=-3pt},
    boxed title style={colframe=black, colback=white}
}

\usepackage{pifont}
\definecolor{lightgreen}{RGB}{120,200,120}
\definecolor{lightred}{RGB}{230,120,120}
\definecolor{lightgray}{gray}{0.92}
\definecolor{cotapp}{RGB}{231,239,250}
\definecolor{refgray}{gray}{0.45}
\newcommand{\cmark}{\textcolor{lightgreen}{\ding{51}}}
\newcommand{\xmark}{\textcolor{lightred}{\ding{55}}}

\begin{document}

\title{CoTA++: Understanding the Repeat Curse in dMLLMs from an Information Flow Perspective}

\author{Qiyan~Zhao,~\IEEEmembership{Member,~IEEE},
        Xiaofeng~Zhang,
        Xu-Yao~Zhang,~\IEEEmembership{Senior Member,~IEEE},
        Shijie~Han,
        Zhepeng~Li,
        Shunzhi~Zhu,
        and Da-Han~Wang,~\IEEEmembership{Senior Member,~IEEE}

\thanks{Q.~Zhao, X.-Y.~Zhang, S.~Zhu and D.-H.~Wang are with the State Key Laboratory of Multimodal Artificial Intelligence Systems, Institute of Automation, Chinese Academy of Sciences, Beijing 100190, China.}%
\thanks{X.~Zhang is with Shanghai Jiao Tong University, Shanghai 200240, China.}%
\thanks{S.~Han and Z.~Li are with the Faculty of Business and Economics, The University of Hong Kong, Hong Kong, China.}%
\thanks{Q.~Zhao and X.-Y.~Zhang are the School of Artificial Intelligence, University of Chinese Academy of Sciences (UCAS), Beijing 100049, China}%
\thanks{Corresponding authors: \mbox{D.-H.~Wang} (e-mail: wangdh@xmut.edu.cn) and \mbox{X.-Y.~Zhang} (e-mail: xyz@nlpr.ia.ac.cn).}%
\thanks{An earlier version of this work appeared at ICLR\,2026~\cite{cota}.}%
\thanks{Code is publicly available at \protect\url{https://github.com/ErikZ719/CoTA}.}}

\markboth{IEEE Transactions on Pattern Analysis and Machine Intelligence,~Vol.~XX, No.~X, 202X}%
{Zhao \MakeLowercase{\textit{et al.}}: CoTA++: Understanding the Repeat Curse in dMLLMs}

\maketitle

\begin{abstract}
Diffusion-based multimodal large language models (dMLLMs) decode by iterative denoising with bidirectional attention, which precludes the standard autoregressive key--value cache. Practical deployments therefore rely on approximate caching that reuses intermediate states across denoising steps. Such caching, however, often makes a response repeat one token at adjacent positions in a run that breaks the sentence, a failure we term the \emph{Repeat Curse}. Tracing the context-token information flow reveals three disruptions. Within a step, caching disperses the attention that anchors a position on nearby context tokens. Across steps, similarity-based refresh is not staleness-aware, so some positions are decoded against stale key--value states. Across layers, the deep-layer entropy of the context fails to converge where repetition arises. We accordingly propose CoTA++, a training-free, plug-and-play framework that counteracts each disruption where it originates: Context-Token Attention Re-anchoring re-forms the attention routing against the current context, Decode-Aware Refresh reserves part of the cache's refresh budget for the positions about to be committed, and Context-Token Entropy-Guided Voting discounts the confidence of a position by the deep-layer entropy of its context, which reorders the unmasking. The three components act at non-overlapping stages of a decoding step and compose freely with existing dMLLMs and caches. Across three dMLLMs and two caching backends, CoTA++ reduces the sample repetition rate of long responses from up to 99.6\% to at most 17.0\%, recovers most of the accuracy lost to caching on nine benchmarks, and keeps the speedup of caching at under 15\% overhead, making approximate caching reliable for dMLLMs.
\end{abstract}


\section{Introduction}
\label{sec:intro}

Diffusion language models (dLLMs)~\cite{llada, llada1.5, dream2025, bie2025llada2} have recently been extended with a vision encoder and vision-instruction tuning~\cite{llava1.5} into diffusion-based multimodal large language models (dMLLMs)~\cite{lladav, mmada, lavida, li2026lavida, zeng2025diffusionvl, ai2026llada2, xin2025lumina}, which now rival their autoregressive counterparts~\cite{llama3, qwen2.5} on multimodal benchmarks. They decode by iterative denoising: starting from a fully masked response, each step predicts all masked positions in parallel and commits the most confident ones. This parallelism promises efficient decoding, but in practice each step recomputes the whole sequence. Bidirectional attention lets every update perturb the key--value (KV) states of all tokens, so the exact KV cache of autoregressive decoding does not apply~\cite{xu2026fixing, chang2026d3tom} and inference remains slow. Recent work instead reuses intermediate states across adjacent steps, an \emph{approximate} caching guided by cross-step feature similarity~\cite{dLLM-Cache, SlowFast, Fast-dLLM}, which yields large speedups and has become the de facto acceleration recipe for dMLLMs~\cite{lladav, mmada, lavida}.

\begin{figure}[t]
\centerline{\includegraphics[width=8.8cm]{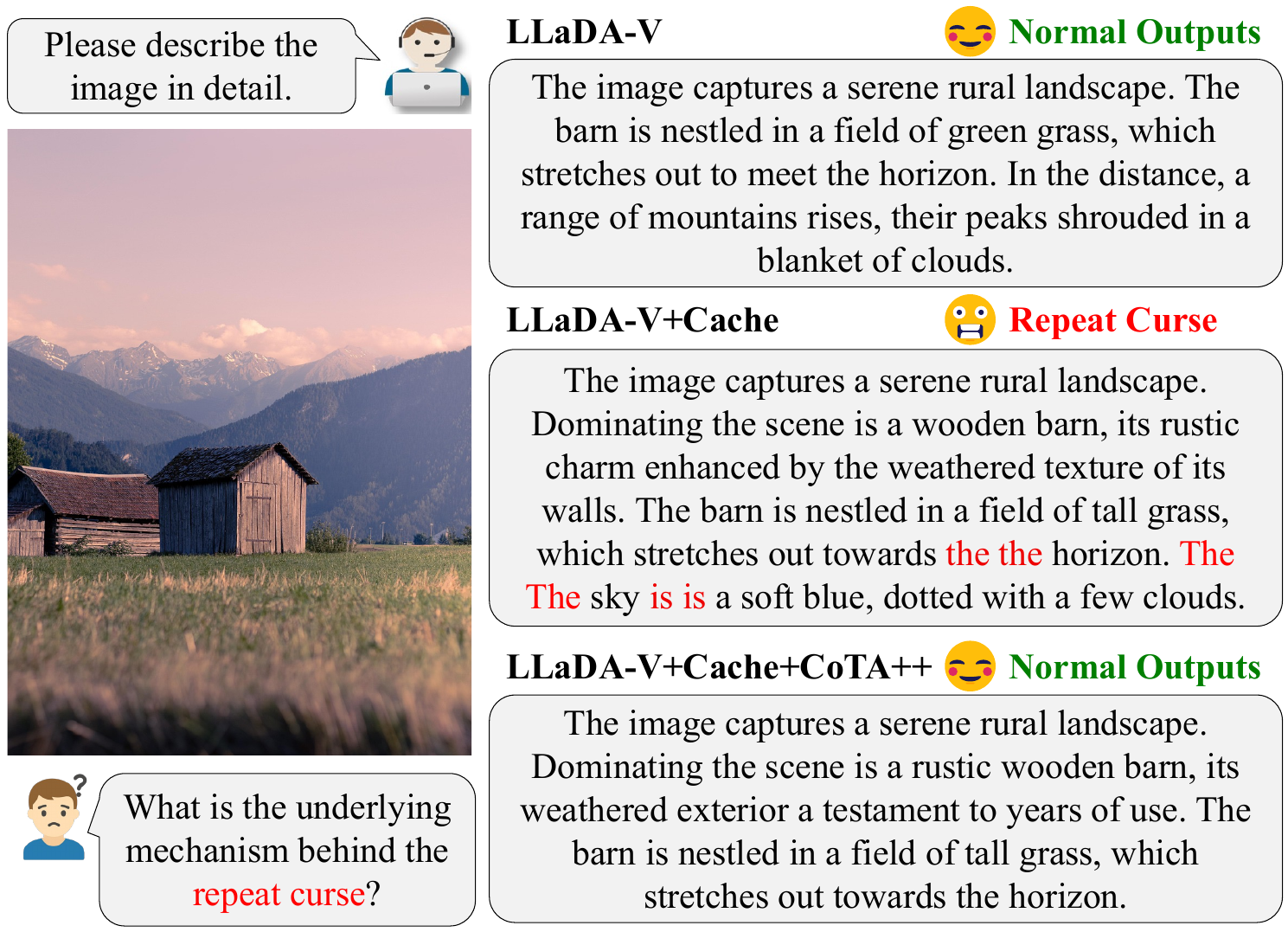}}
\caption{\textbf{Motivation.} LLaDA-V~\cite{lladav} responses to the same query: vanilla decoding is normal (top); dLLM-Cache~\cite{dLLM-Cache} induces the Repeat Curse (repetitions in red, middle); CoTA++ removes them while retaining the acceleration (bottom).}
\label{fig1-motivation}
\vspace{-0.4cm}
\end{figure}

\begin{figure*}[t]
\centering
\vspace{-0.1cm}
\includegraphics[width=2\columnwidth]{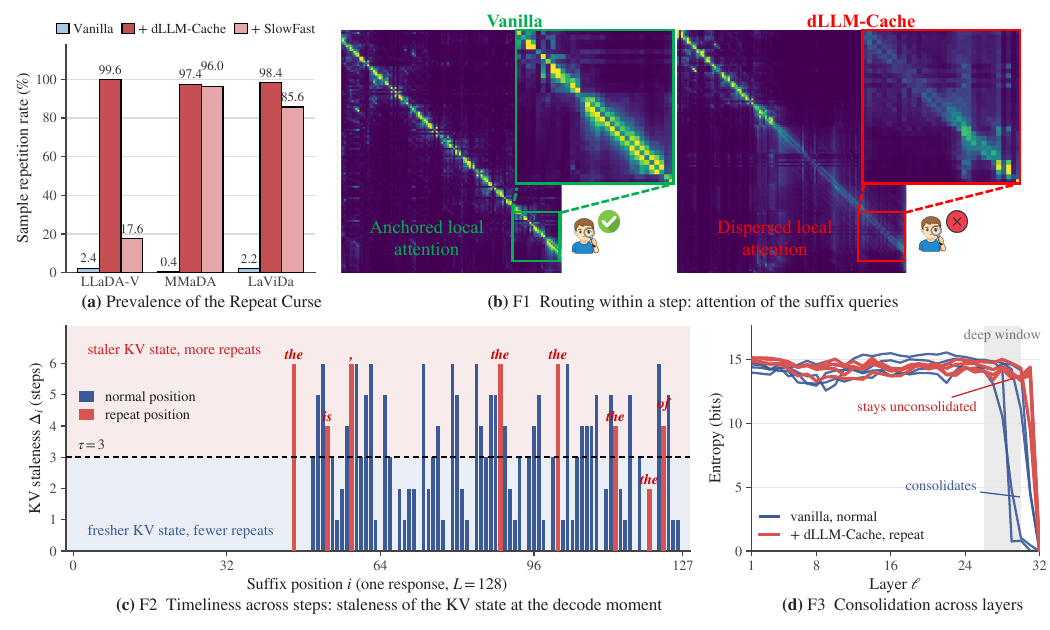}
\vspace{-0.3cm}
\caption{\textbf{The Repeat Curse and its three information-flow signatures.} \emph{(a)} Sample repetition rate on $500$ MS-COCO captions, $L{=}512$. \emph{(b)--(d)} LLaDA-V, $L{=}128$, vanilla against dLLM-Cache. \emph{(b)} Within-step attention of suffix queries: caching disperses local anchoring (F1). \emph{(c)} KV staleness at the decode moment along one response: most repeats (red) decode stale (F2). \emph{(d)} Cross-layer entropy after generation at that response's repeat positions and at normal positions: repeats fail to consolidate (F3).}
\label{fig2-finding1-finding2}
\vspace{-0.35cm}
\end{figure*}

This recipe has a severe side effect. Under caching, a response often repeats one token at adjacent positions, a run that breaks the sentence (Fig.~\ref{fig1-motivation}). We term this failure the \emph{Repeat Curse}. Measured as the share of responses that contain such a run (SRR, \S\ref{sec:moti_metric}) on $500$ MS-COCO~\cite{Lin2014MicrosoftCC} captions, it affects up to $99.6\%$ of long responses and arises across dMLLMs (LLaDA-V~\cite{lladav}, MMaDA~\cite{mmada}, LaViDa~\cite{lavida}) and caching backends (dLLM-Cache~\cite{dLLM-Cache}, SlowFast~\cite{SlowFast}; Fig.~\ref{fig2-finding1-finding2}(a)). Its model- and backend-agnostic nature points to a mechanistic disruption rather than a sampling artifact. The question this paper answers is therefore: \emph{why does approximate caching cause the Repeat Curse in dMLLMs?}

\begin{figure}[t]
\centerline{\includegraphics[width=\columnwidth]{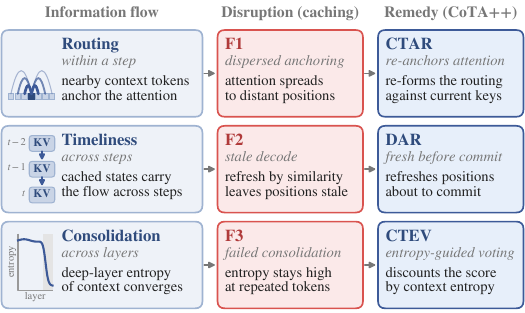}}
\vspace{-3pt}
\caption{\textbf{Overview.} Caching disrupts the information flow of context tokens on the three axes of dMLLM inference, and CoTA++ counteracts each disruption where it originates.}
\label{fig:overview}
\vspace{-0.5cm}
\end{figure}

This question calls for \emph{information-flow} interpretability~\cite{Yu2024UnveilingAH, Wang2023LabelWA, fastv}, which has already served to diagnose abnormal generation~\cite{Yona2025InterpretingTR, zhang2025shallow}. The inference of a dMLLM is indexed by token, step and layer, so information flows along exactly three axes. Along each axis, we trace the flow of the \emph{context tokens}, the tokens around a position being decoded, in the vanilla model and in its cached variant. The result is the thesis of this paper (Fig.~\ref{fig:overview}): the Repeat Curse is the convergent symptom of three distinct disruptions. \textbf{(F1) Routing.} Within a step, a position draws its attention mainly from nearby context tokens, which anchor locally coherent generation. Caching disperses this anchoring where repetition arises (Fig.~\ref{fig2-finding1-finding2}(b)). \textbf{(F2) Timeliness.} Across steps, a position reads its context tokens through cached KV states. The cache refreshes by feature similarity rather than by staleness, so some positions are committed on a view of their context that is several steps old, and repetition concentrates on them (Fig.~\ref{fig2-finding1-finding2}(c)). \textbf{(F3) Consolidation.} Across layers, the deep-layer entropy of context tokens normally converges as the model consolidates its prediction. Under caching it fails to converge at precisely the tokens that repeat, to the end of generation (Fig.~\ref{fig2-finding1-finding2}(d)).

Each disruption prescribes its own remedy, and \textbf{CoTA++}\footnote{The name originates from our conference finding: \textbf{Co}ntext \textbf{T}okens are \textbf{A}nchors~\cite{cota}.} applies the three in one training-free, plug-and-play decoding loop. \textbf{Context-Token Attention Re-anchoring (CTAR)} re-forms the attention of a position against the current context, restoring the anchoring that caching disperses (against~F1). \textbf{Decode-Aware Refresh (DAR)} reserves part of the cache's refresh budget for the positions closest to commitment, so that tokens are not committed against stale states (against~F2). \textbf{Context-Token Entropy-Guided Voting (CTEV)} discounts the confidence of a candidate by the deep-layer entropy of its context, so that the resulting \emph{voting score} favors candidates in consolidated contexts (against~F3). They act on different objects at different moments of a step, the attention output, the refresh set and the voting score, and therefore compose freely with each other and with existing dMLLMs and caches.

\smallskip\noindent\textbf{Extension over the conference version.}
This paper extends our ICLR\,2026 work~\cite{cota}, which introduced the Repeat Curse, identified context tokens as anchors and proposed CTAR\footnote{CTAR was named CTAE in the conference version~\cite{cota}.} and CTEV. The cross-step axis and F2 are new, DAR follows from them, and the study grows to three dMLLMs, two caching backends, three lengths and nine benchmarks, with per-component efficiency and design analyses.

\noindent Our contributions are three-fold:
\begin{enumerate}
\item We characterize the \textbf{Repeat Curse}~\cite{cota} at scale and explain it: tracing the context-token information flow along the token, step and layer axes localizes the failure to attention routing within a step, cache timeliness across steps and entropy consolidation across layers, and controlled comparisons separate the three.
\item We derive \textbf{CoTA++} from this diagnosis: a training-free, plug-and-play framework whose components CTAR, DAR and CTEV counter each disruption where it originates and compose with existing dMLLMs and caches.
\item Across three dMLLMs and two caching backends, CoTA++ reduces the sample repetition rate of long-form generation from up to $99.6\%$ to at most $17.0\%$, recovers most of the accuracy lost to caching on nine benchmarks, and adds under $15\%$ latency overhead.
\end{enumerate}

\section{Related Work}

\subsection{Diffusion Language Models and dMLLMs}

Following the success of diffusion models in image generation~\cite{SDXL, esser2024scaling}, diffusion has been brought to language modeling in continuous~\cite{zhang2025target, gong2023diffuseq} and discrete~\cite{zheng2025masked, ou2025your, nie2025scaling} forms. The masked discrete branch has proved the most scalable: recent dLLMs~\cite{llada, llada1.5, dream2025, xie2025dream, zhu2025llada, bie2025llada2} reach 8B parameters and beyond, with performance comparable to autoregressive LLMs~\cite{llama3, qwen2.5}. Pairing a dLLM with a vision encoder~\cite{Tschannen2025SigLIP2M, Qwen2.5-VL} and a projector, and training it with vision-instruction tuning~\cite{llava1.5}, yields dMLLMs~\cite{lladav, mmada, lavida, li2026lavida, zeng2025diffusionvl, ai2026llada2, 1, 2, 3, 4}. Our study concerns the property that defines this class: inference unrolls many denoising passes over the entire sequence, which makes caching both indispensable and, as we show, hazardous.

\subsection{Inference Acceleration for Diffusion Multimodal LLMs}

For dMLLMs, inference remains the principal deployment bottleneck: every denoising step recomputes the attention and hidden states of the entire multimodal sequence~\cite{lladav, mmada, lavida}. The KV cache of autoregressive decoding~\cite{MLSYS2023_c4be71ab} is exact because causal attention never changes past states. Bidirectional attention perturbs all states at every update, so acceleration resorts to \emph{approximate} caching, which trades an approximation error for the reuse of states across adjacent steps~\cite{dLLM-Cache}.

Existing designs differ in what they reuse and when they refresh. Prefix-KV caching~\cite{lavida} caches the prompt states once, kept exact by its prefix attention mask, and recomputes the whole suffix at every step, which D$^3$ToM~\cite{chang2026d3tom} extends with adaptive visual-token merging. Fast-dLLM~\cite{Fast-dLLM} caches the committed blocks and recomputes from the current block onward. dLLM-Cache~\cite{dLLM-Cache} caches per position, pairing a long refresh interval for the slowly varying prompt with a short one for the suffix and recomputing at every step only the suffix positions whose features change most. Others refine the schedule or the store: SlowFast~\cite{SlowFast} separates cautious exploration from fast parallel decoding, and MaskKV~\cite{MaskKV} evicts cache entries to curb memory. Further variants continue to appear~\cite{FlashDLM, SeedDiffusion, TimeFeature, dkvcache}.

\begin{table}[t]
\centering
\caption{\textbf{Positioning of CoTA++} among dMLLM caching methods. Only CoTA++ makes its refresh decode-aware.}
\label{tab:comparison}
\renewcommand{\arraystretch}{1.15}
\setlength{\tabcolsep}{3.5pt}
\resizebox{\columnwidth}{!}{%
\begin{tabular}{lcccc}
\toprule
\textbf{Method} &
\begin{tabular}[c]{@{}c@{}}Cache\\granularity\end{tabular} &
\begin{tabular}[c]{@{}c@{}}Refresh\\criterion\end{tabular} &
\begin{tabular}[c]{@{}c@{}}Decode-\\aware\end{tabular} &
\begin{tabular}[c]{@{}c@{}}Intervention\\level\end{tabular} \\
\midrule
Fast-dLLM~\cite{Fast-dLLM}   & block        & block boundary          & \xmark & schedule \\
Prefix-KV~\cite{lavida}      & prefix only  & none (exact)            & n/a    & schedule \\
dLLM-Cache~\cite{dLLM-Cache} & per position & periodic $+$ similarity & \xmark & schedule \\
SlowFast~\cite{SlowFast}     & per position & periodic $+$ similarity & \xmark & schedule $+$ decoding \\
dkv-cache~\cite{dkvcache}    & per position & delayed commit          & \xmark & schedule \\
CoTA~\cite{cota}             & plug-in      & inherited               & \xmark & attention $+$ decoding \\
\rowcolor{lightgray}\textbf{CoTA++ (Ours)} & plug-in & inherited $+$ commitment rank & \cmark & attention $+$ cache $+$ decoding \\
\bottomrule
\end{tabular}}
\vspace{-0.2cm}
\end{table}

Across these designs, the refresh schedule is a global, per-step budget: none consults the decoder's ranking of the positions it is about to commit, and none intervenes in the flow that consumes those states. They are also evaluated mainly by task accuracy, which overlooks repetition that leaves the final answer intact, rather than on the open-ended long-form generation in which the approximation error compounds (\S\ref{exp:repeat}). CoTA++ (Table~\ref{tab:comparison}) is a training-free plug-in that inherits the backend's cache, makes its refresh decode-aware, and intervenes where the cached states are consumed.

\subsection{Information-Flow Interpretability}
\enlargethispage{-1\baselineskip}

Information-flow analysis~\cite{abnar2020quantifying, ferrando2024information, zhang2025cross} offers an interpretable lens on the internal mechanisms of black-box models. Its instruments include attention-map analysis~\cite{qiu2026gated, song2026sparse, DPad}, entropy-based readouts~\cite{Yao2025UnderstandingTR, nguyen2025beyond}, saliency attribution~\cite{zhang2026fragile, Wang2023LabelWA}, Grad-CAM~\cite{zhang2025redundancy, li2025token} and massive-value analyses~\cite{gan2026unleashing, gan2026massive}; our diagnosis uses the first two. In autoregressive (M)LLMs, such analyses have traced abnormal outputs to their internal mechanisms, including hallucination~\cite{OPERA, zhaoMCA} and text degeneration~\cite{Yona2025InterpretingTR, olsson2022context}. For dLLMs, interpretability is only beginning to emerge, with recent analyses of dynamic attention sinks~\cite{rulli2026attention, zhang2026one} and attribution methods for dMLLMs~\cite{zuo2026diffusion}. We bring this lens to cached dMLLMs and trace the flow of context tokens along all three axes of their inference.

\subsection{Token Repetition in Generation}

Early research on token repetition and degeneration focused on n-gram penalties~\cite{zhu2023penalty}, decoding strategies~\cite{su2022contrastive, fu2021theoretical} and training-time objectives~\cite{lagutinmplicit, DITTO}, and later work extended the study to modern LLMs~\cite{li2023repetition, Frustratingly, LZPenalty}. Most of them mitigate repetition without explaining it. The exceptions analyze autoregressive models, attributing repetition to latent features~\cite{Yao2025UnderstandingTR} or to specific neurons~\cite{WangLL0S024}. In dMLLMs, our conference version~\cite{cota} was, to our knowledge, the first to identify and analyze repetitive generation, tracing it to approximate caching. A more recent study~\cite{hong2026mitigating} shows that cache-free decoding also repeats under aggressive step reduction, and attributes this to mask-prior drift. This paper deepens the caching line: the three-axis information-flow analysis of \S\ref{moti} explains the Repeat Curse, and a mitigation follows for each axis.

\section{Preliminaries}\label{prelim}

This section formalizes the building blocks on which our analysis and method rest: the dMLLM architecture (\S\ref{dMLLMsArchitecture}), iterative denoising inference (\S\ref{infer}) and approximate KV caching (\S\ref{cache}). Throughout the paper we adopt LLaDA-V~\cite{lladav} and dLLM-Cache~\cite{dLLM-Cache} as the running examples of a representative dMLLM and caching backend.

\subsection{Architecture of dMLLMs}\label{dMLLMsArchitecture}

A dMLLM $\mathcal{F}$ such as LLaDA-V comprises three components: a pretrained vision encoder $\mathcal{F}_v$, a diffusion-based language model (dLLM) $\mathcal{F}_t$ and a projector $f$ that maps visual features into the text embedding space. Given an image $I_v$ and an instruction prompt $I_p$ (e.g., \textit{``Please describe the image in detail''}), the image is encoded and projected into visual tokens $\mathcal{S}_{v} = f(\mathcal{F}_v(I_v))$ and the instruction is tokenized into instruction tokens $\mathcal{S}_{p}$. Their concatenation $\{\mathcal{S}_{v}, \mathcal{S}_{p}\}$ constitutes the \emph{prefix}. A fully masked \emph{suffix} $\mathcal{S}_{m} = \{[\text{MASK}]_{i}\}_{i=1}^{M}$ of maximum response length $M$, written $L$ when we vary it in the experiments, is then appended, and the complete sequence $\mathcal{S} = \{\mathcal{S}_{v}, \mathcal{S}_{p}, \mathcal{S}_{m}\}$ is fed to $\mathcal{F}_t$ for iterative denoising. All tokens share the same hidden dimensionality $d$.

\subsection{Iterative Denoising Inference}\label{infer}

Let $\mathcal{S}^{t} = \{\mathcal{S}_{v}^{t}, \mathcal{S}_{p}^{t}, \mathcal{S}_{m}^{t}\}$ denote the sequence state before decoding step $t \in \{0, 1, \ldots, N-1\}$, initialized as $\mathcal{S}^{0} = \mathcal{S}$ with its suffix entirely masked. The prefix tokens are fixed throughout ($\mathcal{S}_{v}^{t} = \mathcal{S}_{v}$, $\mathcal{S}_{p}^{t} = \mathcal{S}_{p}$), and only the suffix evolves. Over the $N$ steps, the model progressively commits tokens at masked positions, and the suffix of the final state, $\mathcal{S}_{m}^{N}$, is the generated response. At each step $t$, the model estimates, for every masked position $i$, a categorical distribution over the vocabulary $V$, and the corresponding greedy prediction is
\begin{equation}
\hat{\mathcal{S}}^{t}_{(i)} = \arg\max_{v \in V}\, p_{\theta}\!\left(\mathcal{S}_{(i)} = v \,\big|\, \mathcal{S}^{t}\right), \quad i \in \mathcal{I}_{\text{mask}}^{t},
\label{pred}
\end{equation}
where $\mathcal{S}_{(i)}$ denotes the token at suffix position $i$ and $\mathcal{I}_{\text{mask}}^{t}$ the set of masked suffix positions in $\mathcal{S}^{t}$. Since a single parallel pass yields unreliable predictions at many positions, dMLLMs commit only the most confident predictions at each step and re-predict the rest, a strategy known as \emph{confidence-based remasking}. The confidence score of each masked position is
\begin{equation}
c^{t}_{(i)} = p_{\theta}\!\left(\mathcal{S}_{(i)} = \hat{\mathcal{S}}^{t}_{(i)} \,\big|\, \mathcal{S}^{t}\right), \quad i \in \mathcal{I}_{\text{mask}}^{t}.
\label{ci}
\end{equation}
The model selects the $k$ positions with the highest $c^{t}_{(i)}$ ($k = M/N$ under the uniform schedule used throughout), forming the update set $\mathcal{I}_{\text{update}}^{t}$, and produces
\begin{equation}
\mathcal{S}^{t+1}_{(i)} =
\begin{cases}
\hat{\mathcal{S}}^{t}_{(i)}, & \text{if } i \in \mathcal{I}_{\text{update}}^{t}, \\
\mathcal{S}^{t}_{(i)},        & \text{otherwise},
\end{cases}
\label{update}
\end{equation}
with $\mathcal{I}_{\text{mask}}^{t+1} = \mathcal{I}_{\text{mask}}^{t} \setminus \mathcal{I}_{\text{update}}^{t}$. We drop the step superscript of $c^{t}_{(i)}$ where the step is clear from the context.

\subsection{Approximate KV Caching for dMLLMs}\label{cache}

Because dMLLM attention is bidirectional rather than causal, no token's KV states are guaranteed to stay fixed across denoising steps, and practical schemes therefore rely on \emph{approximate} caching. They share a common starting point: the features of prefix tokens $\{\mathcal{S}_{v}, \mathcal{S}_{p}\}$ and of most suffix tokens vary slowly across adjacent steps~\cite{dLLM-Cache, SlowFast}. dLLM-Cache, our running backend, accordingly organizes per-step recomputation by token type:
\begin{equation}
\text{Recompute at step } t:\quad
\begin{cases}
\{\mathcal{S}_{v}^{t}, \mathcal{S}_{p}^{t}\},  & \text{if } t \bmod \mathcal{E}_p = 0, \\[3pt]
\mathcal{S}_{m}^{t},                            & \text{if } t \bmod \mathcal{E}_s = 0, \\[3pt]
\mathcal{S}_{(i)}^{t},                          & \text{if } i \in \mathcal{B}_{\alpha}^{t},
\end{cases}
\label{eq:cache}
\end{equation}
where prefix states are refreshed every $\mathcal{E}_p$ steps and the entire suffix every $\mathcal{E}_s$ steps. At all other steps, $\mathcal{B}_{\alpha}^{t}$ collects the $\lfloor \alpha M \rfloor$ suffix positions with the lowest cross-step similarity $s_{(i)}^{t}$, the cosine between position $i$'s current and cached features, so only the tokens whose features have drifted most are recomputed, and the rest are reused.\footnote{Unless otherwise stated, hyperparameters are fixed to $\alpha = 0.25$, $\mathcal{E}_p = 25$, and $\mathcal{E}_s = 7$ for fair comparison.}

\section{Motivation and Analysis}\label{moti}

This section analyzes how approximate KV caching (\S\ref{cache}) triggers the Repeat Curse. After defining the metrics (\S\ref{sec:moti_metric}), it takes the three axes of Section~\ref{sec:intro} in turn, \emph{routing} within a step (F1, \S\ref{sec:moti_f1}), \emph{timeliness} across steps (F2, \S\ref{sec:moti_f2}) and \emph{consolidation} across layers (F3, \S\ref{sec:moti_f3}), and closes with a joint account that separates the three disruptions (\S\ref{sec:moti_joint}). Each finding is summarized in a box of the same form: observation, mechanism, evidence and implication.

\subsection{Quantifying the Repeat Curse}\label{sec:moti_metric}

\noindent\textbf{Repeat-Curse metrics.} To quantify the token repetition illustrated in Fig.~\ref{fig1-motivation}, we introduce the \emph{Adjacent Repetition Rate} (ARR), computed on the content tokens $y = (y_{1},\dots,y_{|y|})$ of a generated response (the decoded suffix $\mathcal{S}_{m}^{N}$ of \S\ref{infer}, truncated at its first end-of-text token):
\begin{equation}
\text{ARR} = \frac{1}{|y|-1} \sum_{i=1}^{|y|-1} \mathbf{1}\!\left( y_{i} = y_{i+1} \right),
\end{equation}
where $\mathbf{1}(\cdot)$ is the indicator function. The second token of each adjacent identical pair marks a \emph{repeat position}, the criterion followed by every position-level analysis in this section. The \emph{Sample Repetition Rate} (SRR) measures the prevalence of repetition over a test set of $n$ responses:
\begin{equation}
\text{SRR} = n_{\text{rep}} / n,
\end{equation}
where $n_{\text{rep}}$ counts the responses with $\text{ARR}>0$. Three additional metrics characterize the \emph{severity} of repetition: with $\{r_k\}$ collecting the lengths of all maximal runs of identical tokens ($r_k \ge 2$) over the $n$ responses, we report the maximum, average and 95th-percentile repetition lengths $\text{MRL} = \max_k r_k$, $\text{ARL} = \operatorname{mean}_k\, r_k$ and $\text{95pRL} = \operatorname{Quantile}_{0.95}\{r_k\}$. Classical degeneration metrics measure windowed or $n$-gram recurrence~\cite{Welleck2020Neural, su2022contrastive}. ARR instead targets the runs of identical tokens that dominate the Repeat Curse.

\begin{figure}[t]
\vspace{-0.1cm}
\centerline{\includegraphics[width=\columnwidth]{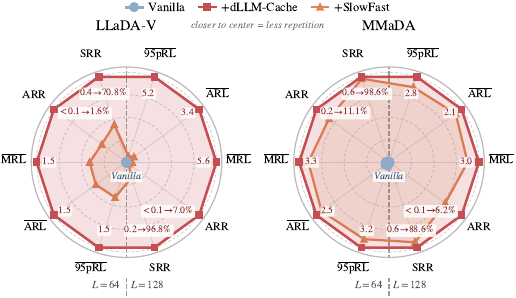}}
\vspace{-0.1cm}
\caption{\textbf{Radar charts of the Repeat Curse.} Near zero without caching, every metric surges with it (labels: vanilla${\to}$cached).}
\label{radar}
\vspace{-0.2cm}
\end{figure}
\noindent\textbf{Severity across models, backends and settings.} Figure~\ref{radar} visualizes the Repeat Curse on LLaDA-V and MMaDA under two caching backends at $L{=}64$ and $L{=}128$, on detailed captioning over $500$ MS-COCO images~\cite{Lin2014MicrosoftCC} (full values in Table~\ref{tabrepeat}). Each axis is scaled by its value under dLLM-Cache, printed on the outer ring, and overbars denote run-length metrics averaged per response. Without a cache both models are essentially repetition-free, with an SRR of at most $0.6\%$. dLLM-Cache raises the SRR of LLaDA-V to $70.8\%$ at $L{=}64$ and $96.8\%$ at $L{=}128$, a $484\times$ increase at the longer length, and that of MMaDA to $98.6\%$ and $88.6\%$. ARR follows, rising from below $0.2\%$ to between $1.6\%$ and $11.1\%$. SlowFast is markedly milder on LLaDA-V, yet still lifts its SRR by nearly two orders of magnitude, to $31.2\%$ and $17.4\%$, and on MMaDA it is barely milder than dLLM-Cache ($96.4\%$ and $83.0\%$). Which backend is milder therefore depends on the model, and neither avoids the curse.

\newcommand{\swp}[2]{\begin{tabular}[c]{@{}c@{}}{\scriptsize\textcolor{gray}{#1}}\\[-2pt]#2\end{tabular}}

\smallskip
\noindent\textbf{Localizing the responsible cache component.} To identify \emph{which} cache component drives the failure, Table~\ref{ablation_cache} sweeps the three design choices of Eq.~(\ref{eq:cache}), namely the prefix interval $\mathcal{E}_p$, the suffix interval $\mathcal{E}_s$ and the similarity fraction $\alpha$, and contrasts dLLM-Cache with a prefix-only reuse policy~\cite{lavida}. Varying $\mathcal{E}_p$ barely moves the SRR, which stays between $96.8\%$ and $99.0\%$. Stretching $\mathcal{E}_s$ from $1$ to $7$ drives it from $0.6\%$ to $96.8\%$, and lowering $\alpha$ from $1$ to $0$ drives it from $0.4\%$ to $99.8\%$. The two ends agree: $\mathcal{E}_s{=}1$, which is the prefix-only policy, and $\alpha{=}1$ both recompute every suffix position at every step, and both return the SRR to the uncached level. The Repeat Curse is thus driven mainly by the approximate caching of suffix tokens, and the suffix is the segment that our subsequent analysis and method target.

\begin{figure*}[t]
\centerline{\includegraphics[width=17cm]{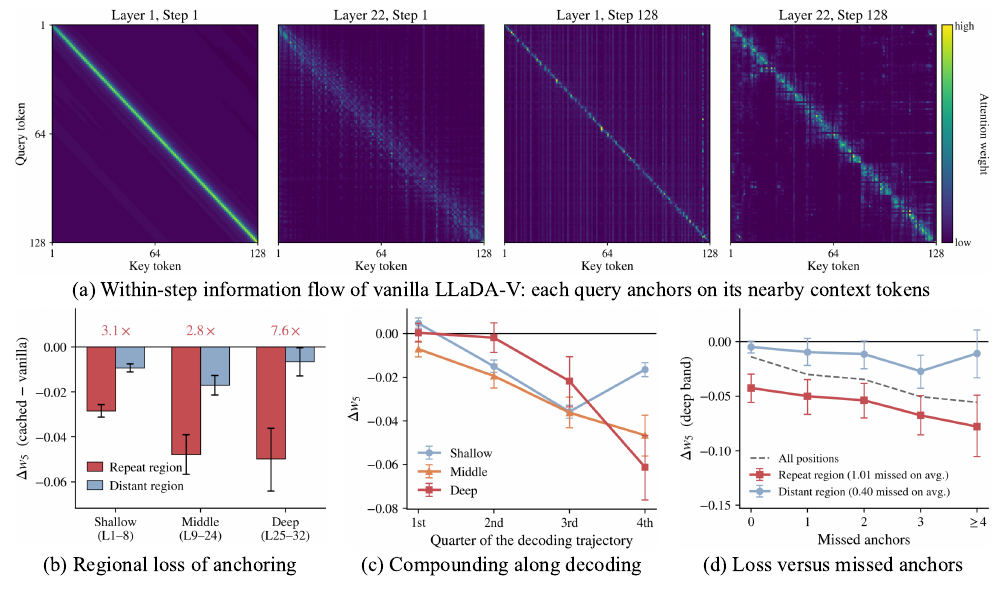}}
\caption{\textbf{Context tokens are anchors, and caching disperses them (Finding F1).} \emph{(a)} Attention of vanilla LLaDA-V (brighter is higher). \emph{(b)--(d)} Change in local anchoring under dLLM-Cache on $94$ MS-COCO images, $\Delta w_{5}$ (cached $-$ vanilla), where $w_{5}$ is the share of a position's suffix attention within ${\pm}5$ positions. Repeat regions: repeat positions ${\pm}5$. Distant regions: all other positions. Numbers in \emph{(b)}: ratio of the two losses. Error bars: $95\%$ intervals over images.}
\label{fig:f1}
\vspace{-6pt}
\end{figure*}

\begin{table}[!t]
\vspace{-3pt}
\centering
\small
\caption{\textbf{Ablation of dLLM-Cache components} on LLaDA-V ($L{=}128$; SRR, \%). Each row sweeps one component with the others at their defaults (shaded).}
\label{ablation_cache}
\setlength{\tabcolsep}{5pt}
\renewcommand{\arraystretch}{1.15}
\begin{tabular}{@{}lcccc@{}}
\toprule
\textbf{Component sweep} & \multicolumn{4}{c}{\textbf{SRR (\%)}$\downarrow$} \\
\midrule
Prefix interval $\mathcal{E}_p$ & \swp{$1$}{99.0} & \swp{$5$}{97.2} & \swp{$15$}{97.6} & \cellcolor{gray!12}\swp{$25$}{96.8} \\
Suffix interval $\mathcal{E}_s$ & \swp{$1$}{0.6} & \swp{$3$}{68.4} & \swp{$5$}{90.8} & \cellcolor{gray!12}\swp{$7$}{96.8} \\
Similarity fraction $\alpha$ & \swp{$0$}{99.8} & \cellcolor{gray!12}\swp{$0.25$}{96.8} & \swp{$0.75$}{25.2} & \swp{$1$}{0.4} \\
Reuse policy & \swp{prefix-only}{0.6} & \cellcolor{gray!12}\swp{dLLM-Cache}{96.8} & --- & --- \\
\bottomrule
\end{tabular}
\vspace{-4pt}
\end{table}
\subsection{Routing Within a Step: Dispersed Anchoring (F1)}\label{sec:moti_f1}

We begin with the within-step \emph{routing} axis and visualize the attention of the suffix queries. In the uncached model, the attention maps show a pronounced diagonal that persists across layers and decoding steps (Fig.~\ref{fig:f1}(a)): queries concentrate on their \emph{context tokens}, the tokens at neighboring positions. The pattern resembles the anchor tokens of autoregressive LLMs, such as attention sinks~\cite{xiao2024streamingllm} and label words~\cite{Wang2023LabelWA}, which absorb a large share of attention and relay information. The anchors here, however, are \emph{relative}: under bidirectional attention every query has its own, set by distance rather than by token identity. Approximate caching disperses this anchoring where repetition arises (Fig.~\ref{fig2-finding1-finding2}(b)): attention there moves away from the context tokens and spreads over distant positions.

\smallskip
\noindent\textbf{Quantifying the dispersion.} Local anchoring is scored by $w_{5}$, the share of a position's suffix attention that falls within ${\pm}5$ positions. We compare the cached with the vanilla run at each position's own decoding step, on the $94$ of $100$ MS-COCO images that repeat under dLLM-Cache. The loss of anchoring is \emph{regional} (Fig.~\ref{fig:f1}(b)): repeat regions (repeat positions ${\pm}5$) lose $2.8$ to $7.6$ times as much as distant regions, most in the deep band. It also \emph{compounds along decoding} (Fig.~\ref{fig:f1}(c)), growing with every quarter of the trajectory in the middle and deep bands. Both hold across images (all $p<10^{-6}$). Yet even in repeat regions the loss averages about a tenth of the vanilla anchoring mass: anchoring is weakened where repetition emerges, not abolished.

\smallskip
\noindent\textbf{Dispersion grows with the anchors a position has missed.} The cache reuses the routing of a position until it is formed again, so a context token within ${\pm}5$ committed in the meantime is an anchor that this routing predates, a \emph{missed anchor}. The deep-band loss grows with the number of missed anchors (Fig.~\ref{fig:f1}(d)), and repeat regions miss more than twice as many as distant ones ($1.01$ against $0.40$ per position). Both observations hold across images (both \mbox{$p<10^{-10}$}).

\vspace{-17pt}
\begin{findingbox}[Finding F1]
\textbf{F1: Context tokens act as information-flow anchors.} \mbox{\textup{\textbf{Observation.}}} Neighboring context tokens anchor the within-step attention routing. Caching disperses this anchoring where repetition arises. \textup{\textbf{Mechanism.}} A cached routing misses the anchors committed since it was formed. \textup{\textbf{Evidence.}} The loss is regional, compounds along decoding and grows with the missed anchors (Fig.~\ref{fig:f1}). \textup{\textbf{Implication.}} The routing must be re-formed against the current context tokens, which is what CTAR does (\S\ref{sec:ctar}).
\end{findingbox}

\subsection{Timeliness Across Steps: Staleness-Driven Fracture (F2)}\label{sec:moti_f2}

We next turn to the cross-step \emph{timeliness} axis. Across steps, each prediction reads the evolving sequence state through the KV states of the surrounding positions. Once these states are cached and reused (\S\ref{cache}), the cache carries the flow across steps, and its refresh policy decides how timely that flow is. Approximate caches such as dLLM-Cache refresh by feature similarity, a criterion that is not \emph{staleness-aware}: a minority of suffix positions go unrefreshed for several consecutive steps and are decoded against outdated KV states. The analysis below formalizes this lag and traces it to the repeat positions.

\smallskip
\noindent\textbf{Cache staleness $\Delta_i^{t}$.} Let $R^{t}\subseteq\{1,\ldots,M\}$ be the set of suffix positions recomputed at step~$t$ under the cache's refresh policy. Refresh decisions are made per layer. $R^{t}$ is their union across layers, and all statistics below are computed from it. The staleness of position $i$ at step $t$,
\begin{equation}
\Delta_i^{t} \;=\; t \;-\; \max\{\, t' \le t : i \in R^{t'} \,\},
\label{eq:staleness}
\end{equation}
counts the decoding steps since the position was last recomputed, so that $\Delta_i^{t}{=}0$ exactly when $i \in R^{t}$. At staleness $\Delta$, a position is thus decoded against KV states that reflect $\mathcal{S}^{t-\Delta}$ or an earlier state. Because $R^{t}$ is a union, $\Delta$ lower-bounds the staleness of each layer, and the bound is tight in over $99\%$ of the layers at the decode moment (see the supplementary material). Vanilla dMLLMs recompute every position at every step ($\Delta \equiv 0$), so each prediction is conditioned only on the current state. Approximate caching breaks this guarantee wherever $\Delta_i^{t}>0$.

\smallskip
\begin{figure*}[t]
\centerline{\includegraphics[width=17cm]{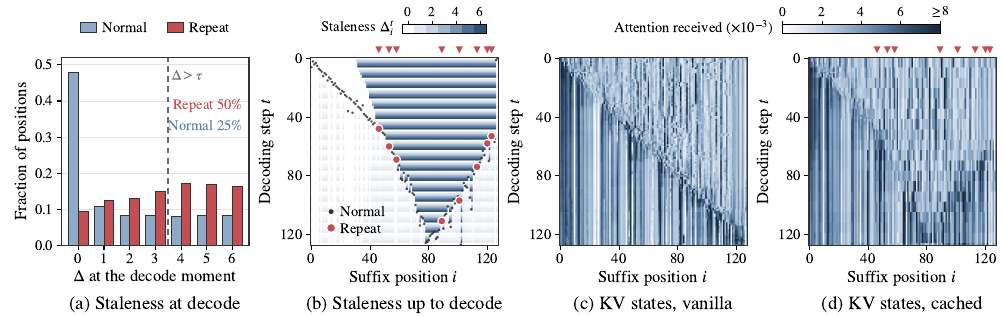}}
\caption{\textbf{Cache staleness concentrates on repeat positions (Finding F2).} \emph{(a)} Staleness at the decode moment on $94$ images. \emph{(b)}--\emph{(d)} Reference sample, on shared axes: \emph{(b)} staleness of each position up to its decode (faded afterwards), \emph{(c)} and \emph{(d)} how each KV state is read at layer $22$ (attention received, averaged over heads and queries). Triangles: repeat positions.}
\label{recom}
\vspace{-9pt}
\end{figure*}
\begin{table}[t]
\centering
\footnotesize
\caption{\textbf{Staleness, refresh and repetition} on $94$ images. $\Delta{>}\tau$: share decoded beyond $\tau{=}3$. Images: where repeat exceeds normal. $p$: Wilcoxon signed-rank test over images.}
\vspace{-2.5pt}
\label{tab:staleness_quartile}
\vskip 0.03in
\setlength{\tabcolsep}{0pt}
\renewcommand{\arraystretch}{1.15}
\begin{tabular*}{\columnwidth}{@{\extracolsep{\fill}}lcccccc@{}}
\toprule
 & \multicolumn{2}{c}{Positions} & \multicolumn{2}{c}{$\Delta{>}\tau$} & & \\
\cmidrule{2-3}\cmidrule{4-5}
(a) Quarter (steps) & Repeat & Normal & Repeat & Normal & Images & $p$ \\
\midrule
1st (0--31) & $15$ & $2{,}899$ & $0.20$ & $0.02$ & $3/13$ & $0.54$ \\
2nd (32--63) & $110$ & $2{,}898$ & $0.63$ & $0.19$ & $43/51$ & $7.8{\times}10^{-9}$ \\
3rd (64--95) & $337$ & $2{,}671$ & $0.47$ & $0.40$ & $51/70$ & $2.7{\times}10^{-5}$ \\
4th (96--127) & $484$ & $2{,}524$ & $0.50$ & $0.42$ & $64/86$ & $5.1{\times}10^{-6}$ \\
All & $946$ & $10{,}992$ & $0.50$ & $0.25$ & $85/94$ & $3.4{\times}10^{-15}$ \\
\end{tabular*}\par\nointerlineskip\vspace{1.5pt}
\begin{tabular*}{\columnwidth}{@{\extracolsep{\fill}}lccccc@{}}
\midrule
(b) Recomputed per step (\%) & 1st & 2nd & 3rd & 4th & All \\
\midrule
Committed at the previous step & $100$ & $100$ & $100$ & $100$ & $100$ \\
Committed, all & $60.8$ & $59.0$ & $40.3$ & $28.7$ & $40.0$ \\
Still-masked & $20.0$ & $5.1$ & ${<}0.1$ & ${<}0.1$ & $10.2$ \\
\end{tabular*}\par\nointerlineskip\vspace{1.5pt}
\begin{tabular*}{\columnwidth}{@{\extracolsep{\fill}}lccccccc@{}}
\midrule
(c) Steps after a full refresh & $0$ & $1$ & $2$ & $3$ & $4$ & $5$ & $6$ \\
\midrule
Repeat positions (\%) & $4.9$ & $6.6$ & $7.5$ & $8.6$ & $9.7$ & $9.4$ & $9.1$ \\
\multicolumn{8}{@{}l}{Steps $4$--$6$ against $0$: \ $83/94$ images, \ $p{=}1.2{\times}10^{-13}$} \\
\bottomrule
\end{tabular*}
\vspace{-11pt}
\end{table}
\noindent\textbf{Cache staleness concentrates on repeat positions.} Staleness is read at each position's \emph{decode moment}, the step at which it is committed. On the same $94$ images, normal positions concentrate at low staleness (median $1$) and repeat positions in the stale tail (median $4$, Fig.~\ref{recom}(a)). Half of the repeat positions are decoded beyond $\tau{=}3$, the midpoint of the staleness range, against a quarter of the normal ones ($p<10^{-14}$ across images). Staleness and repetition both grow along the trajectory, yet the contrast is not merely one of lateness: in every quarter after the first, repeat positions remain staler than normal ones (all $p\le2.7\times10^{-5}$, Table~\ref{tab:staleness_quartile}), most markedly in the second, where repetition sets in ($0.63$ against $0.19$).

\begin{figure}[t]
\centerline{\includegraphics[width=\columnwidth]{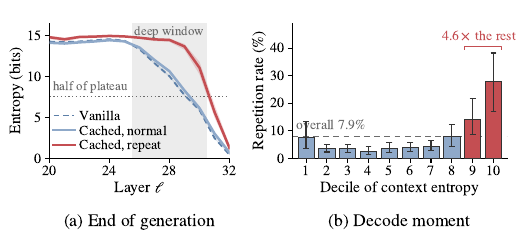}}
\caption{\textbf{Consolidation fails at repeat positions (Finding F3).} \emph{(a)} Entropy per layer (shaded: deep window). \emph{(b)} Repetition rate by decile of the deep-window entropy of the committed neighbors within ${\pm}5$. Intervals: $95\%$, over $94$ images.}
\label{fig:f3quant}
\vspace{-10pt}
\end{figure}

\smallskip
\noindent\textbf{How the cross-step flow loses timeliness.} dLLM-Cache recomputes the whole suffix every seventh step and, in between, only the quarter whose features changed most. Committing a position replaces its mask token with a word, so it is always recomputed at the next step, whereas a still-masked position's input is unchanged. The selection thus favors committed positions ($40\%$ against $10\%$ per step) and, in the second half of decoding, reaches a still-masked one in under $0.1\%$ of the steps (Table~\ref{tab:staleness_quartile}(b)). The KV states of the positions about to be committed thus grow stale until the next full refresh (Fig.~\ref{recom}\mbox{(b)--(d)}). Repetition follows this cycle independently of lateness, nearly doubling from the step of a full refresh to four steps later ($4.9\%$ against $9.7\%$, $p{=}1.2\times10^{-13}$, Table~\ref{tab:staleness_quartile}(c)).

\vspace{-17pt}
\begin{findingbox}[Finding F2]
\textbf{F2: Cache refresh is not staleness-aware.} \textup{\textbf{Observation.}} Cached KV states carry the information flow of the context tokens across steps. Repetition concentrates where those states are stale, beyond what lateness explains. \textup{\textbf{Mechanism.}} Refresh by similarity favors committed positions, so the KV states of those about to be committed grow stale until the next full refresh. \textup{\textbf{Evidence.}} Repeat positions concentrate in the stale tail in every quarter after the first, and repetition rises along the refresh cycle (Fig.~\ref{recom}, Table~\ref{tab:staleness_quartile}). \textup{\textbf{Implication.}} The flow into a position must be refreshed before it is committed, which DAR enforces (\S\ref{sec:dar}).
\end{findingbox}

\subsection{Consolidation Across Layers: Failed Consolidation (F3)}\label{sec:moti_f3}

Finally, we turn to the cross-layer \emph{consolidation} axis: whether the information carried by the flow condenses, layer by layer, into a confident prediction. We measure it by the entropy of the per-layer prediction distribution. Reading the layer-$\ell$ hidden state of a token through the output head of the model (a logit-lens readout~\cite{EntropyLens}) yields a distribution $p^{(\ell)}_{v}$ over the vocabulary $V$, whose entropy in bits is
\begin{equation}
E^{(\ell)} = -\sum_{v \in V} p^{(\ell)}_{v} \log_2 p^{(\ell)}_{v}.
\label{eq:entropy}
\end{equation}
Low deep-layer entropy indicates a consolidated prediction.

\smallskip
\noindent\textbf{Repeat positions remain unconsolidated.} In a normal response, the entropy of a token stays on a plateau through the shallow and middle layers and converges over the deep window $\mathcal{W}{=}\{26,\dots,30\}$, as the model aggregates cross-layer information into a confident prediction~\cite{EntropyLens} (Fig.~\ref{fig2-finding1-finding2}(d), blue). Under approximate caching, normal positions keep this profile (Fig.~\ref{fig:f3quant}(a)), whereas at repeat positions the entropy stays on the plateau throughout the deep window and collapses only in the last two layers (Fig.~\ref{fig2-finding1-finding2}(d), red). The contrast holds on the $94$ images (Fig.~\ref{fig:f3quant}(a)): at the end of generation, $79\%$ of the repeat positions have not shed half of their plateau entropy before layer $31$, against $36\%$ of the normal ones ($p<10^{-14}$ across images, Table~\ref{tab:f3_stats}). A repeated token thus remains unconsolidated even after generation has finished.

\begin{table}[t]
\centering
\footnotesize
\caption{\textbf{Consolidation and repetition} on $94$ images. Top fifth: highest context entropy. Images: where the first value exceeds the second. $p$: Wilcoxon signed-rank test over images.}
\vspace{-2.5pt}
\label{tab:f3_stats}
\vskip 0.03in
\setlength{\tabcolsep}{2.5pt}
\renewcommand{\arraystretch}{1.15}
\begin{tabular*}{\columnwidth}{@{\extracolsep{\fill}}lcccc@{}}
\toprule
(a) End of generation & Repeat & Normal & Images & $p$ \\
\midrule
Deep-window entropy (bits) & $13.7$ & $10.1$ & $92/94$ & $4.6{\times}10^{-17}$ \\
Halved not before layer $31$ (\%) & $79$ & $36$ & $89/94$ & $2.0{\times}10^{-15}$ \\
\midrule
(b) Decode moment & Top fifth & Rest & Images & $p$ \\
\midrule
Repetition rate (\%) & $21.2$ & $4.6$ & $66/94$ & $1.5{\times}10^{-7}$ \\
Share of all repetitions (\%) & $54$ & $46$ & & \\
\bottomrule
\end{tabular*}
\vspace{-6pt}
\end{table}

\smallskip
\noindent\textbf{Failed consolidation predicts repetition.} The failure shows in the local context before a repetition is committed. We score each position by its \emph{context entropy}, the entropy of its committed neighbors within ${\pm}5$ averaged over the deep window and read at its decode moment, and bin the positions by decile of this score (Fig.~\ref{fig:f3quant}(b)). The repetition rate stays at or below the overall rate ($7.9\%$) in the first eight deciles and rises steeply in the last two. The fifth of positions with the highest context entropy repeats $4.6$ times as often as the rest and holds more than half of all repetitions ($p<10^{-6}$ across images, Table~\ref{tab:f3_stats}). An unconsolidated local context thus signals that a repetition is about to be committed.

\vspace{-17pt}
\begin{findingbox}[Finding F3]
\textbf{F3: The information flow fails to consolidate at repeat positions.} \textup{\textbf{Observation.}} A normal token's deep-layer entropy typically converges, whereas a repeated token's stays high through the deep window. \textup{\textbf{Mechanism.}} A repetition is committed from an unconsolidated context. \textup{\textbf{Evidence.}} Most repeat positions keep over half their entropy through the deep window, and the top fifth by context entropy holds over half of all repetitions (Fig.~\ref{fig:f3quant}, Table~\ref{tab:f3_stats}). \textup{\textbf{Implication.}} The context entropy should discount the confidence, which CTEV does in its voting score (\S\ref{sec:ctev}).
\end{findingbox}

\subsection{Joint Diagnosis}\label{sec:moti_joint}

The three findings describe one failure along the three axes of the information flow (Fig.~\ref{fig:overview}), each with its own object: routing in the attention (F1), timeliness in the KV states (F2) and consolidation in the deep-layer entropy (F3). They are not redundant: staleness accounts for most of the excess anchoring loss of repeat regions in the shallow band ($81\%$) but for almost none in the deep band ($5\%$), and failed consolidation persists at equal staleness. They also compound: tokens committed with dispersed routing, stale KV states and an unconsolidated context repeat at $55\%$, against $2.4\%$ with none of them (see the supplementary material). Each disruption thus calls for an intervention on its own object, which Section~\ref{methodsec} derives.

\begin{figure*}[t]
\centerline{\includegraphics[width=17cm]{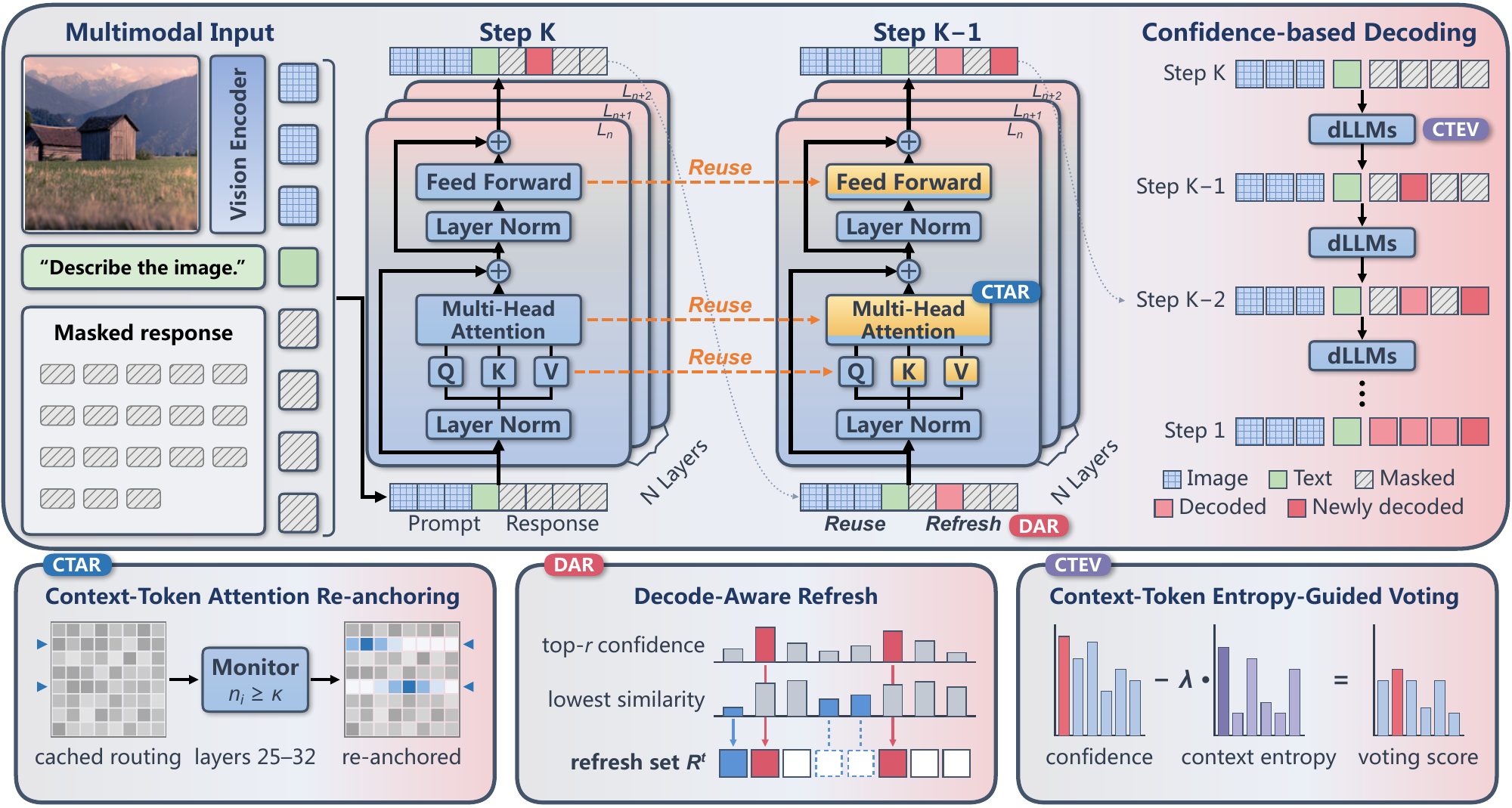}}
\caption{\textbf{The CoTA++ framework.} \emph{Top}: a cached dMLLM reuses the feed-forward, attention and KV states of Step~$K$ at Step~$K{-}1$ and unmasks the response in order of confidence. Badges mark where each component acts. \emph{Bottom}: each component acts on its own object. CTAR re-anchors the \emph{attention} of the positions its monitor selects, DAR refreshes the \emph{KV states} of the positions about to commit, and CTEV discounts the confidence by the deep-layer \emph{entropy} of the context.}
\label{fig:framework}
\vspace{-10pt}
\end{figure*}

\section{Method: The CoTA++ Framework}\label{methodsec}

\textbf{CoTA++} (Fig.~\ref{fig:framework}) is a training-free, plug-and-play framework that turns the three implications of Section~\ref{moti} into three components, one per disruption. Its design follows one principle: intervene where the disruption arises, on the object it corrupts. \textbf{CTAR} re-forms the \emph{routing} of the within-step flow (F1, \S\ref{sec:ctar}), \textbf{DAR} restores the \emph{timeliness} of the cross-step flow (F2, \S\ref{sec:dar}), and \textbf{CTEV} brings the \emph{consolidation} of the cross-layer flow into the order of commitment (F3, \S\ref{sec:ctev}). Section~\ref{sec:algo} places them in one loop and examines their cost.

\subsection{Context-Token Attention Re-anchoring (CTAR)}\label{sec:ctar}

For a position it does not recompute, the cache replays the attention \emph{output} stored at an earlier step, whose routing predates every anchor committed since (\S\ref{sec:moti_f1}). The dispersion thus lies in the routing itself, so CTAR re-forms the routing against the current context. Let $h_i^{(\ell)}$ denote the hidden state that suffix position~$i$ carries into layer~$\ell$ at the current step, and $K^{(\ell)}$ and $V^{(\ell)}$ the keys and values the cache currently holds. CTAR recomputes the attention distribution of the position against these keys,
\begin{equation}
q_i^{(\ell)} = W_{Q}^{(\ell)}\,\mathrm{LN}\bigl(h_i^{(\ell)}\bigr),
\quad
A_{i}^{(\ell)} = \operatorname{softmax}\bigl( q_i^{(\ell)} K^{(\ell)\top} / \sqrt{d_h} \,\bigr),
\label{eq:ctar_route}
\end{equation}
and reads its attention output through this distribution,
\begin{equation}
o_{i}^{(\ell)} \;=\; W_{O}^{(\ell)} \bigl( A_{i}^{(\ell)} V^{(\ell)} \bigr),
\label{eq:ctar_out}
\end{equation}
where $W_{Q}^{(\ell)}$ and $W_{O}^{(\ell)}$ are the query and output projections of the layer, $\mathrm{LN}$ its input normalization and $d_h$ the dimension of one attention head. The result replaces the stored output of position~$i$, in the current pass and in the cache. The position is thus routed against the context as it now stands.

Re-anchoring every position at every step would recompute the attention of the whole suffix. The dispersion grows with the anchors a position has missed (Fig.~\ref{fig:f1}(d)), so CTAR re-anchors a position only once it has missed one, that is, once one of its context tokens has been committed (\S\ref{sec:moti_f1}). A \emph{context-anchor monitor} reads this from the decoding schedule alone: for every masked position it counts the context tokens within ${\pm}w$ committed since its last re-anchoring, or since decoding began, and at each step CTAR evaluates Eqs.~(\ref{eq:ctar_route})--(\ref{eq:ctar_out}) for the set $\mathcal{T}^{t}$ of masked positions whose count has reached a threshold~$\kappa$. With $\kappa{=}1$, a position is re-anchored as soon as one of its context tokens is committed. The count ignores the recomputations of the backend, which keeps the monitor independent of whichever refresh policy is in use.

CTAR acts in the deep band $\mathcal{D}{=}\{25,\dots,32\}$, where the dispersion is most concentrated on repeat regions (Fig.~\ref{fig:f1}(b)). It adds no learned parameter and no operation of its own: the routing is re-formed by the attention of the model, while the keys, values and feed-forward outputs stay cached. Every masked position thus stays anchored to its context tokens in the deep band as they are committed (\textbf{against~F1}).

\subsection{Decode-Aware Refresh (DAR)}\label{sec:dar}

A similarity-based cache refreshes the positions whose features have changed most, and the positions still masked are rarely among them (Table~\ref{tab:staleness_quartile}(b)). The positions about to be committed are thus decoded on stale KV states (\S\ref{sec:moti_f2}). Timeliness matters at the step of decision: a stale state is inconsequential at a position that stays masked and decisive at one about to be committed. DAR therefore makes the refresh aware of the decoding schedule and reserves part of its budget for the positions about to be committed, which the confidence ranking of the decoder already identifies.

Let $\mathcal{I}^{t}$ be the $r$ masked positions with the highest confidence $c_{(i)}$ of Eq.~(\ref{ci}) at step $t{-}1$. DAR refreshes them at step $t$ and leaves the rest of the refresh set to the cache. With $b$ the number of positions that the cache refreshes at a step and $s^{t}_{(i)}$ the similarity by which it selects them (\S\ref{cache}), the set $R^{t}$ of positions recomputed at step~$t$ (Eq.~(\ref{eq:staleness})) becomes
\begin{equation}
R^{t} \;=\; \mathcal{I}^{t} \,\cup\, \operatorname{Bottom\text{-}}k\bigl(\{\,s^{t}_{(i)} : i \notin \mathcal{I}^{t}\,\}\bigr),
\label{eq:dar_set}
\end{equation}
with $\operatorname{Bottom\text{-}}k$ the $k{=}b{-}r$ positions of lowest similarity.

The refresh set keeps its size $b$, so DAR spends no more than the cache and only redirects what it spends. It leaves the selection rule of the backend in place and has a single hyperparameter, $r$. By Eq.~(\ref{eq:staleness}), the reserved positions have staleness $\Delta{=}0$ at step~$t$. The positions about to be committed are thus decoded on fresh KV states (\textbf{against~F2}).

\subsection{Context-Token Entropy-Guided Voting (CTEV)}\label{sec:ctev}

A dMLLM commits a token on its confidence alone, which reports how peaked the prediction is, not whether the context behind it has consolidated. Under caching the two diverge where repetition arises (\S\ref{sec:moti_f3}). F3 located the signature of repetition in the context a position commits from, so CTEV brings the context entropy into the decision. For a masked position~$i$, it averages the entropy $E^{(\ell)}$ of Eq.~(\ref{eq:entropy}) over the deep window $\mathcal{W}{=}\{26,\dots,30\}$ and over the context tokens already committed,
\begin{equation}
\bar{E}_{\text{ctx}}(i) \;=\; \frac{1}{|\mathcal{N}(i)|\,|\mathcal{W}|}\sum_{j \in \mathcal{N}(i)}\,\sum_{\ell \in \mathcal{W}} E^{(\ell)}(j),
\label{eq:ctev_ctx}
\end{equation}
where $\mathcal{N}(i) = \{\, j \notin \mathcal{I}_{\text{mask}}^{t} : 0 < |j-i| \le w \,\}$ collects the committed context tokens within ${\pm}w$, and $\bar{E}_{\text{ctx}}(i){=}0$ if no context token has been committed yet.

This is the quantity whose top fifth repeats $4.6$ times as often as the rest in Fig.~\ref{fig:f3quant}(b). CTEV subtracts it from the confidence,
\begin{equation}
\text{Score}(i) \;=\; c_{(i)} \;-\; \lambda\, \frac{\bar{E}_{\text{ctx}}(i)}{\log_2 |V|},
\qquad \lambda > 0,
\label{eq:ctev_score}
\end{equation}
where dividing by $\log_2|V|$, the largest possible entropy, puts the penalty on the confidence scale. The resulting \emph{voting score} replaces $c_{(i)}$ in the top-$k$ selection of Eq.~(\ref{update}). Each committed context token thus votes through its deep-layer entropy, and
a candidate in an unconsolidated context yields its turn to one of comparable confidence in a consolidated context.

The penalty is graded, so CTEV never forbids a token and only changes the order in which tokens are committed. It adds no learned parameter, reuses the readout of \S\ref{sec:moti_f3} and leaves the cache untouched. Positions whose context has converged are thus committed ahead of the rest (\textbf{against~F3}).

\begin{algorithm}[t]
\small
\caption{CoTA++ Inference Loop}
\label{alg:cotapp}
\begin{algorithmic}[1]
\REQUIRE Initial sequence $\mathcal{S}$; steps $N$; band $\mathcal{D}$, window $w$ and threshold $\kappa$ (CTAR); reserved slots $r$ (DAR); penalty $\lambda$ (CTEV)
\ENSURE Decoded response $\mathcal{S}_m^{N}$
\STATE $\mathcal{D} \gets \{25,\dots,32\}$; $n_i \gets 0$ for all $i$; $c^{-1} \gets \mathbf{0}$
\FOR{$t = 0, 1, \ldots, N-1$}
  \STATE $\mathcal{T}^{t} \gets \{ i \in \mathcal{I}_{\text{mask}}^{t} : n_i \ge \kappa \}$; reset $n_i$ for $i \in \mathcal{T}^{t}$ \hfill \COMMENT{monitor}
  \STATE $\mathcal{I}^{t} \gets \operatorname{Top\text{-}}r(c^{t-1})$; assemble $R^{t}$ via Eq.~(\ref{eq:dar_set}) \hfill \COMMENT{DAR}
  \STATE Forward pass, recomputing the cached state at $R^{t}$; for $\ell \in \mathcal{D}$ and $i \in \mathcal{T}^{t}$ re-derive $q^{(\ell)}_i$ and $A^{(\ell)}_i$ via Eq.~(\ref{eq:ctar_route}) and re-project via Eq.~(\ref{eq:ctar_out}) \hfill \COMMENT{CTAR}
  \STATE Compute $c_{(i)}$ via Eq.~(\ref{ci}) and $\bar{E}_{\text{ctx}}(i)$ via Eq.~(\ref{eq:ctev_ctx}) \hfill \COMMENT{CTEV}
  \STATE $\text{Score}(i) \gets c_{(i)} - \lambda\, \bar{E}_{\text{ctx}}(i)/\log_2|V|$ \hfill \COMMENT{CTEV}
  \STATE Select the top-$k$ positions by $\text{Score}(i)$ and unmask them via Eq.~(\ref{update}); $n_j \gets n_j + 1$ for every $j$ within ${\pm}w$ of a committed position \hfill \COMMENT{monitor}
\ENDFOR
\RETURN $\mathcal{S}_m^{N}$
\end{algorithmic}
\end{algorithm}

\begin{table}[t]
\centering
\caption{\textbf{Per-step cost} of the three components. $\mathcal{D}$: re-anchored layer band; $\mathcal{T}^{t}$: positions selected by the monitor; $\mathcal{W}$: deep window of the entropy readout; $M$: suffix length; $S$: sequence length; $d$: hidden size; $|V|$: vocabulary size.}
\label{tab:complexity}
\renewcommand{\arraystretch}{1.15}
\setlength{\tabcolsep}{4pt}
\footnotesize
\begin{tabular}{lll}
\toprule
\textbf{Component} & \textbf{Intervention level} & \textbf{Extra cost per step} \\
\midrule
CTAR & Attention routing, in-pass      & $O(|\mathcal{D}|\,|\mathcal{T}^{t}|\,(S{+}d)\,d)$ \\
DAR  & Refresh set, at the step boundary & $O(M)$ index ops \\
CTEV & Voting score, after the pass      & $O(|\mathcal{W}|\,M d\,|V|)$ \\
\bottomrule
\end{tabular}
\vspace{-2mm}
\end{table}

\begin{table*}[t]
\centering
\providecommand{\na}{\textcolor{refgray}{--}}
\caption{\textbf{Repeat Curse across dMLLMs and caching backends} on the $500$ COCO images at three response lengths $L$, under the instruction ``Please describe the image in detail.'', on content tokens (\S\ref{setDetails}). ARR and SRR are in percent. MRL, ARL and 95pRL are the maximum, mean and 95th percentile of the lengths of all repeated runs in the $500$ responses. Avg.: mean over the three lengths. $\downarrow$: lower is better. Bold: best within a backend group. Shaded: CoTA++.}
\label{tabrepeat}
\scriptsize
\setlength{\tabcolsep}{2.6pt}
\renewcommand{\arraystretch}{1.12}
\begin{tabular}{@{}c l ccccc c ccccc c ccccc c cc}
\toprule
& & \multicolumn{5}{c}{$L{=}512$} & & \multicolumn{5}{c}{$L{=}128$} & & \multicolumn{5}{c}{$L{=}64$} & & \multicolumn{2}{c}{Avg.} \\
\cmidrule(lr){3-7} \cmidrule(lr){9-13} \cmidrule(lr){15-19} \cmidrule(l){21-22}
Model & Method & ARR$\downarrow$ & SRR$\downarrow$ & MRL$\downarrow$ & ARL$\downarrow$ & 95pRL$\downarrow$ &  & ARR$\downarrow$ & SRR$\downarrow$ & MRL$\downarrow$ & ARL$\downarrow$ & 95pRL$\downarrow$ &  & ARR$\downarrow$ & SRR$\downarrow$ & MRL$\downarrow$ & ARL$\downarrow$ & 95pRL$\downarrow$ &  & ARR$\downarrow$ & SRR$\downarrow$ \\
\midrule
 & \textcolor{refgray}{Vanilla} & 0.01 & 2.4 & 3 & 2.07 & 2.6 &  & 0.00 & 0.2 & 2 & 2.00 & 2.0 &  & 0.01 & 0.4 & 2 & 2.00 & 2.0 &  & 0.01 & 1.0 \\
\cmidrule(l){2-22}
 & dLLM-Cache~\cite{dLLM-Cache} & 14.70 & 99.6 & 455 & 5.90 & 3.0 &  & 7.01 & 96.8 & 73 & 2.88 & 3.0 &  & 1.61 & 70.8 & 38 & 2.12 & \textbf{2.0} &  & 7.77 & 89.1 \\
 & \quad$+$\,CoTA~\cite{cota} & 0.17 & 47.4 & \textbf{3} & \textbf{2.00} & \textbf{2.0} &  & 0.10 & 11.4 & \textbf{2} & \textbf{2.00} & \textbf{2.0} &  & 0.04 & 2.4 & \textbf{2} & \textbf{2.00} & \textbf{2.0} &  & 0.10 & 20.4 \\
 & \cellcolor{cotapp}\quad$+$\,\textbf{CoTA++} & \cellcolor{cotapp}\textbf{0.03} & \cellcolor{cotapp}\textbf{8.2} & \cellcolor{cotapp}\textbf{3} & \cellcolor{cotapp}2.12 & \cellcolor{cotapp}3.0 & \cellcolor{cotapp} & \cellcolor{cotapp}\textbf{0.02} & \cellcolor{cotapp}\textbf{2.4} & \cellcolor{cotapp}\textbf{2} & \cellcolor{cotapp}\textbf{2.00} & \cellcolor{cotapp}\textbf{2.0} & \cellcolor{cotapp} & \cellcolor{cotapp}\textbf{0.01} & \cellcolor{cotapp}\textbf{0.4} & \cellcolor{cotapp}\textbf{2} & \cellcolor{cotapp}\textbf{2.00} & \cellcolor{cotapp}\textbf{2.0} & \cellcolor{cotapp} & \cellcolor{cotapp}\textbf{0.02} & \cellcolor{cotapp}\textbf{3.7} \\
\cmidrule(l){2-22}
 & SlowFast~\cite{SlowFast} & 0.28 & 17.6 & 232 & 2.89 & \textbf{2.0} &  & 0.22 & 17.4 & 3 & 2.03 & \textbf{2.0} &  & 0.55 & 31.2 & \textbf{2} & \textbf{2.00} & \textbf{2.0} &  & 0.35 & 22.1 \\
 & \quad$+$\,CoTA~\cite{cota} & 0.15 & 6.8 & 286 & 8.33 & \textbf{2.0} &  & 0.04 & 3.2 & \textbf{2} & \textbf{2.00} & \textbf{2.0} &  & 0.03 & 1.4 & \textbf{2} & \textbf{2.00} & \textbf{2.0} &  & 0.07 & 3.8 \\
\multirow{-7}{*}{\rotatebox[origin=c]{90}{LLaDA-V~\cite{lladav}}} & \cellcolor{cotapp}\quad$+$\,\textbf{CoTA++} & \cellcolor{cotapp}\textbf{0.02} & \cellcolor{cotapp}\textbf{3.2} & \cellcolor{cotapp}\textbf{3} & \cellcolor{cotapp}\textbf{2.10} & \cellcolor{cotapp}3.0 & \cellcolor{cotapp} & \cellcolor{cotapp}\textbf{0.02} & \cellcolor{cotapp}\textbf{1.6} & \cellcolor{cotapp}\textbf{2} & \cellcolor{cotapp}\textbf{2.00} & \cellcolor{cotapp}\textbf{2.0} & \cellcolor{cotapp} & \cellcolor{cotapp}\textbf{0.01} & \cellcolor{cotapp}\textbf{0.4} & \cellcolor{cotapp}\textbf{2} & \cellcolor{cotapp}\textbf{2.00} & \cellcolor{cotapp}\textbf{2.0} & \cellcolor{cotapp} & \cellcolor{cotapp}\textbf{0.02} & \cellcolor{cotapp}\textbf{1.7} \\
\midrule
 & \textcolor{refgray}{Vanilla} & 0.00 & 0.4 & 2 & 2.00 & 2.0 &  & 0.01 & 0.6 & 2 & 2.00 & 2.0 &  & 0.19 & 0.6 & 57 & 20.33 & 51.5 &  & 0.07 & 0.5 \\
\cmidrule(l){2-22}
 & dLLM-Cache~\cite{dLLM-Cache} & 11.35 & 97.4 & 409 & 3.22 & 5.0 &  & 6.18 & 88.6 & 34 & 2.50 & 4.0 &  & 11.08 & 98.6 & 30 & 2.48 & 4.0 &  & 9.54 & 94.9 \\
 & \quad$+$\,CoTA~\cite{cota} & 0.18 & 14.2 & 7 & 2.12 & 3.0 &  & 0.11 & 9.6 & 3 & 2.02 & \textbf{2.0} &  & 0.25 & 13.0 & 3 & 2.06 & 2.6 &  & 0.18 & 12.3 \\
 & \cellcolor{cotapp}\quad$+$\,\textbf{CoTA++} & \cellcolor{cotapp}\textbf{0.00} & \cellcolor{cotapp}\textbf{0.2} & \cellcolor{cotapp}\textbf{2} & \cellcolor{cotapp}\textbf{2.00} & \cellcolor{cotapp}\textbf{2.0} & \cellcolor{cotapp} & \cellcolor{cotapp}\textbf{0.02} & \cellcolor{cotapp}\textbf{1.6} & \cellcolor{cotapp}\textbf{2} & \cellcolor{cotapp}\textbf{2.00} & \cellcolor{cotapp}\textbf{2.0} & \cellcolor{cotapp} & \cellcolor{cotapp}\textbf{0.02} & \cellcolor{cotapp}\textbf{1.2} & \cellcolor{cotapp}\textbf{2} & \cellcolor{cotapp}\textbf{2.00} & \cellcolor{cotapp}\textbf{2.0} & \cellcolor{cotapp} & \cellcolor{cotapp}\textbf{0.01} & \cellcolor{cotapp}\textbf{1.0} \\
\cmidrule(l){2-22}
 & SlowFast~\cite{SlowFast} & 8.68 & 96.0 & 400 & 2.97 & 5.0 &  & 4.78 & 83.0 & 35 & 2.41 & 4.0 &  & 9.15 & 96.4 & 35 & 2.40 & 4.0 &  & 7.54 & 91.8 \\
 & \quad$+$\,CoTA~\cite{cota} & 0.14 & 12.2 & 3 & 2.03 & \textbf{2.0} &  & 0.15 & 13.8 & \textbf{2} & \textbf{2.00} & \textbf{2.0} &  & 0.19 & 10.4 & 3 & 2.02 & \textbf{2.0} &  & 0.16 & 12.1 \\
\multirow{-7}{*}{\rotatebox[origin=c]{90}{MMaDA~\cite{mmada}}} & \cellcolor{cotapp}\quad$+$\,\textbf{CoTA++} & \cellcolor{cotapp}\textbf{0.01} & \cellcolor{cotapp}\textbf{1.2} & \cellcolor{cotapp}\textbf{2} & \cellcolor{cotapp}\textbf{2.00} & \cellcolor{cotapp}\textbf{2.0} & \cellcolor{cotapp} & \cellcolor{cotapp}\textbf{0.01} & \cellcolor{cotapp}\textbf{0.8} & \cellcolor{cotapp}\textbf{2} & \cellcolor{cotapp}\textbf{2.00} & \cellcolor{cotapp}\textbf{2.0} & \cellcolor{cotapp} & \cellcolor{cotapp}\textbf{0.01} & \cellcolor{cotapp}\textbf{0.4} & \cellcolor{cotapp}\textbf{2} & \cellcolor{cotapp}\textbf{2.00} & \cellcolor{cotapp}\textbf{2.0} & \cellcolor{cotapp} & \cellcolor{cotapp}\textbf{0.01} & \cellcolor{cotapp}\textbf{0.8} \\
\midrule
 & \textcolor{refgray}{Vanilla} & 0.02 & 2.2 & 3 & 2.27 & 3.0 &  & 0.05 & 4.6 & 3 & 2.26 & 3.0 &  & 0.02 & 1.0 & 2 & 2.00 & 2.0 &  & 0.03 & 2.6 \\
\cmidrule(l){2-22}
 & dLLM-Cache~\cite{dLLM-Cache} & 6.62 & 98.4 & 465 & 3.36 & 3.0 &  & 1.56 & 75.6 & 89 & 2.21 & \textbf{2.0} &  & 1.64 & 24.6 & 47 & 2.56 & 3.0 &  & 3.27 & 66.2 \\
 & \quad$+$\,CoTA~\cite{cota} & 0.38 & 43.8 & 3 & 2.02 & \textbf{2.0} &  & 0.16 & 18.4 & \textbf{3} & \textbf{2.02} & \textbf{2.0} &  & 0.04 & 2.2 & 3 & 2.09 & 2.5 &  & 0.19 & 21.5 \\
 & \cellcolor{cotapp}\quad$+$\,\textbf{CoTA++} & \cellcolor{cotapp}\textbf{0.04} & \cellcolor{cotapp}\textbf{6.6} & \cellcolor{cotapp}\textbf{2} & \cellcolor{cotapp}\textbf{2.00} & \cellcolor{cotapp}\textbf{2.0} & \cellcolor{cotapp} & \cellcolor{cotapp}\textbf{0.07} & \cellcolor{cotapp}\textbf{9.2} & \cellcolor{cotapp}\textbf{3} & \cellcolor{cotapp}2.02 & \cellcolor{cotapp}\textbf{2.0} & \cellcolor{cotapp} & \cellcolor{cotapp}\textbf{0.03} & \cellcolor{cotapp}\textbf{1.8} & \cellcolor{cotapp}\textbf{2} & \cellcolor{cotapp}\textbf{2.00} & \cellcolor{cotapp}\textbf{2.0} & \cellcolor{cotapp} & \cellcolor{cotapp}\textbf{0.05} & \cellcolor{cotapp}\textbf{5.9} \\
\cmidrule(l){2-22}
 & SlowFast~\cite{SlowFast} & 5.67 & 85.6 & 428 & 2.72 & 3.0 &  & 1.41 & 79.6 & 41 & 2.10 & \textbf{2.0} &  & 0.69 & 6.4 & 50 & 3.49 & 3.0 &  & 2.59 & 57.2 \\
 & \quad$+$\,CoTA~\cite{cota} & 0.34 & 28.4 & \textbf{3} & 2.02 & \textbf{2.0} &  & 0.20 & 17.8 & 16 & 2.31 & \textbf{2.0} &  & \textbf{0.03} & \textbf{1.4} & 3 & 2.14 & 2.7 &  & 0.19 & 15.9 \\
\multirow{-7}{*}{\rotatebox[origin=c]{90}{LaViDa~\cite{lavida}}} & \cellcolor{cotapp}\quad$+$\,\textbf{CoTA++} & \cellcolor{cotapp}\textbf{0.16} & \cellcolor{cotapp}\textbf{17.0} & \cellcolor{cotapp}\textbf{3} & \cellcolor{cotapp}\textbf{2.01} & \cellcolor{cotapp}\textbf{2.0} & \cellcolor{cotapp} & \cellcolor{cotapp}\textbf{0.07} & \cellcolor{cotapp}\textbf{8.6} & \cellcolor{cotapp}\textbf{3} & \cellcolor{cotapp}\textbf{2.02} & \cellcolor{cotapp}\textbf{2.0} & \cellcolor{cotapp} & \cellcolor{cotapp}0.03 & \cellcolor{cotapp}1.6 & \cellcolor{cotapp}\textbf{2} & \cellcolor{cotapp}\textbf{2.00} & \cellcolor{cotapp}\textbf{2.0} & \cellcolor{cotapp} & \cellcolor{cotapp}\textbf{0.09} & \cellcolor{cotapp}\textbf{9.1} \\
\bottomrule
\end{tabular}
\vspace{-2mm}
\end{table*}

\subsection{Unified Algorithm and Complexity}\label{sec:algo}

Algorithm~\ref{alg:cotapp} places the three components in a single decoding loop. DAR acts at the step boundary, on the refresh set. CTAR acts inside the forward pass, on the attention output of the positions its monitor selects. CTEV acts once the pass is over, on the voting score. They share no state, so that any subset of them can be enabled, and the rest of the loop is left to the backend. The framework requires only what a cached dMLLM already provides, a confidence-ordered unmasking and a cache that exposes its states and its refresh set.

Their costs differ in kind (Table~\ref{tab:complexity}). DAR adds a top-$r$ selection per step, and CTAR a row of attention with its two projections for each selected position in each layer of the band. CTEV adds the deep-window readout, which dominates since each of its $|\mathcal{W}|$ projections spans the vocabulary.

\section{Experiments}\label{exp}

We evaluate CoTA++ on three dMLLMs, each under two caching backends (\S\ref{setDetails}). The experiments address three questions: whether CoTA++ mitigates the Repeat Curse across models, backends and response lengths without a cost in response quality (\S\ref{exp:repeat}); whether it preserves the general multimodal ability of the model, on short answers and on long ones (\S\ref{exp:general}); and what each component contributes, at what cost in latency and memory, and whether it acts on the axis where its finding located the disruption (\S\ref{exp:analysis}).

\subsection{Experimental Setup}\label{setDetails}

\noindent\textbf{Models and caching backends.} We evaluate three dMLLMs, LLaDA-V-8B~\cite{lladav}, MMaDA-8B~\cite{mmada} and LaViDa-8B~\cite{lavida}, each under two approximate caches, dLLM-Cache~\cite{dLLM-Cache} and SlowFast~\cite{SlowFast}. LLaDA-V and LaViDa share the LLaDA-8B-Instruct language tower~\cite{llada} and a two-layer MLP projector, over a siglip2-so400m-patch14-384~\cite{Tschannen2025SigLIP2M} and a siglip-so400m-patch14-384 vision tower respectively. MMaDA~\cite{mmada} is instead a single diffusion model over a joint vocabulary, where an image enters as discrete MAGVIT-v2 tokens. Unless stated otherwise, one token is committed per step, and the caches run at the released dLLM-Cache setting on LLaDA-V ($\mathcal{E}_p{=}25$, $\mathcal{E}_s{=}7$, $\alpha{=}0.25$). Each response is truncated at its first end-of-text token before any metric is computed.

\noindent\textbf{Benchmarks.} The Repeat Curse is measured on the descriptions of $500$ MS-COCO val2014 images~\cite{Lin2014MicrosoftCC}, whose list we release with our code, at response lengths of $64$, $128$ and $512$ tokens. General multimodal ability is measured on nine benchmarks, DocVQA~\cite{mathew2021docvqa}, ChartQA~\cite{masry2022chartqa}, MMStar~\cite{mmstar}, MME~\cite{mme}, SEED-Bench~\cite{li2023seed}, MMBench~\cite{liu2024mmbench}, LLaVA-Bench (in-the-wild)~\cite{llava1.5}, MathVista~\cite{lu2024mathvista} and MathVerse~\cite{zhang2024mathverse}, each evaluated under its released protocol.

\noindent\textbf{Metrics.} Repetition is measured by ARR, SRR, MRL, ARL and 95pRL (\S\ref{sec:moti_metric}). ARR and SRR are averaged over the responses, and the run lengths are pooled over all of them. Since adjacent runs miss repetition at the phrase level, we also report seq-rep-$n$, the share of the $n$-grams of a response that repeat an earlier one, and distinct-$n$, its complement, both read against the uncached model. To verify that repetition is not removed at the cost of quality, every description is rated by a judge on a $1$--$10$ scale for accuracy against the five human captions of the image, level of detail and coherence. The judge is \texttt{gpt-5.6-luna}. Its protocol, the statistical tests and the hardware are given in the supplementary material.

\noindent\textbf{Hyperparameters.} CTAR re-anchors the deep band $\mathcal{D}{=}\{25,\dots,32\}$ with $\kappa{=}1$ and $w{=}5$, DAR reserves $r{=}4$ positions, and CTEV uses $\lambda{=}0.25$ over the deep window $\mathcal{W}{=}\{26,\dots,30\}$, with the same $w$ for its context tokens.

\subsection{Repeat Curse Evaluation}\label{exp:repeat}

\begin{table}[!t]
\centering
\providecommand{\na}{\textcolor{refgray}{--}}
\caption{\textbf{Response quality and length} in the setting of Table~\ref{tabrepeat}. Judge is the mean score of the $500$ descriptions on a $1$--$10$ scale (\S\ref{setDetails}), Len the mean response length in tokens. Bold: best within a backend group. Shaded: CoTA++.}
\label{tabphrase}
\scriptsize
\setlength{\tabcolsep}{2.4pt}
\renewcommand{\arraystretch}{1.0}
\begin{tabular}{@{}c l cc c cc c cc c c}
\toprule
& & \multicolumn{2}{c}{$L{=}512$} & & \multicolumn{2}{c}{$L{=}128$} & & \multicolumn{2}{c}{$L{=}64$} & & \multicolumn{1}{c}{Avg.} \\
\cmidrule(lr){3-4} \cmidrule(lr){6-7} \cmidrule(lr){9-10} \cmidrule(l){12-12}
Model & Method & Judge$\uparrow$ & Len &  & Judge$\uparrow$ & Len &  & Judge$\uparrow$ & Len &  & Judge$\uparrow$ \\
\midrule
 & \textcolor{refgray}{Vanilla} & 3.93 & 475 &  & 4.77 & 127 &  & 4.34 & 63 &  & 4.35 \\
\cmidrule(l){2-12}
 & dLLM-Cache & 3.03 & 509 &  & 3.59 & 127 &  & 4.17 & 63 &  & 3.60 \\
 & \quad$+$\,CoTA & 3.89 & 510 &  & 4.73 & 127 &  & 4.38 & 63 &  & 4.33 \\
 & \cellcolor{cotapp}\quad$+$\,\textbf{CoTA++} & \cellcolor{cotapp}\textbf{4.00} & \cellcolor{cotapp}510 & \cellcolor{cotapp} & \cellcolor{cotapp}\textbf{4.77} & \cellcolor{cotapp}127 & \cellcolor{cotapp} & \cellcolor{cotapp}\textbf{4.40} & \cellcolor{cotapp}63 & \cellcolor{cotapp} & \cellcolor{cotapp}\textbf{4.39} \\
\cmidrule(l){2-12}
 & SlowFast & 4.12 & 254 &  & 4.34 & 99 &  & 3.94 & 56 &  & 4.13 \\
 & \quad$+$\,CoTA & 4.29 & 158 &  & 4.27 & 86 &  & \textbf{4.24} & 57 &  & 4.27 \\
\multirow{-7}{*}{\rotatebox[origin=c]{90}{LLaDA-V}} & \cellcolor{cotapp}\quad$+$\,\textbf{CoTA++} & \cellcolor{cotapp}\textbf{4.30} & \cellcolor{cotapp}238 & \cellcolor{cotapp} & \cellcolor{cotapp}\textbf{4.35} & \cellcolor{cotapp}97 & \cellcolor{cotapp} & \cellcolor{cotapp}4.18 & \cellcolor{cotapp}58 & \cellcolor{cotapp} & \cellcolor{cotapp}\textbf{4.28} \\
\midrule
 & \textcolor{refgray}{Vanilla} & 1.88 & 108 &  & 1.95 & 98 &  & 1.93 & 63 &  & 1.92 \\
\cmidrule(l){2-12}
 & dLLM-Cache & 1.38 & 106 &  & 1.52 & 84 &  & 1.39 & 54 &  & 1.43 \\
 & \quad$+$\,CoTA & \textbf{1.84} & 102 &  & \textbf{1.88} & 92 &  & 1.88 & 61 &  & 1.87 \\
 & \cellcolor{cotapp}\quad$+$\,\textbf{CoTA++} & \cellcolor{cotapp}\textbf{1.84} & \cellcolor{cotapp}103 & \cellcolor{cotapp} & \cellcolor{cotapp}\textbf{1.88} & \cellcolor{cotapp}95 & \cellcolor{cotapp} & \cellcolor{cotapp}\textbf{1.94} & \cellcolor{cotapp}61 & \cellcolor{cotapp} & \cellcolor{cotapp}\textbf{1.89} \\
\cmidrule(l){2-12}
 & SlowFast & 1.41 & 103 &  & 1.48 & 86 &  & 1.44 & 56 &  & 1.44 \\
 & \quad$+$\,CoTA & \textbf{1.89} & 103 &  & 1.87 & 92 &  & \textbf{1.97} & 61 &  & \textbf{1.91} \\
\multirow{-7}{*}{\rotatebox[origin=c]{90}{MMaDA}} & \cellcolor{cotapp}\quad$+$\,\textbf{CoTA++} & \cellcolor{cotapp}1.88 & \cellcolor{cotapp}104 & \cellcolor{cotapp} & \cellcolor{cotapp}\textbf{1.88} & \cellcolor{cotapp}94 & \cellcolor{cotapp} & \cellcolor{cotapp}1.96 & \cellcolor{cotapp}62 & \cellcolor{cotapp} & \cellcolor{cotapp}\textbf{1.91} \\
\midrule
 & \textcolor{refgray}{Vanilla} & 4.60 & 170 &  & 4.76 & 127 &  & 3.74 & 46 &  & 4.36 \\
\cmidrule(l){2-12}
 & dLLM-Cache & 3.07 & 195 &  & 4.28 & 126 &  & 2.51 & 26 &  & 3.28 \\
 & \quad$+$\,CoTA & 4.63 & 162 &  & \textbf{4.78} & 125 &  & \textbf{3.51} & 39 &  & 4.31 \\
 & \cellcolor{cotapp}\quad$+$\,\textbf{CoTA++} & \cellcolor{cotapp}\textbf{4.67} & \cellcolor{cotapp}162 & \cellcolor{cotapp} & \cellcolor{cotapp}\textbf{4.78} & \cellcolor{cotapp}126 & \cellcolor{cotapp} & \cellcolor{cotapp}3.50 & \cellcolor{cotapp}40 & \cellcolor{cotapp} & \cellcolor{cotapp}\textbf{4.32} \\
\cmidrule(l){2-12}
 & SlowFast & 3.01 & 122 &  & 4.17 & 126 &  & 2.21 & 15 &  & 3.13 \\
 & \quad$+$\,CoTA & 3.80 & 91 &  & \textbf{4.74} & 125 &  & 3.38 & 36 &  & 3.97 \\
\multirow{-7}{*}{\rotatebox[origin=c]{90}{LaViDa}} & \cellcolor{cotapp}\quad$+$\,\textbf{CoTA++} & \cellcolor{cotapp}\textbf{3.85} & \cellcolor{cotapp}95 & \cellcolor{cotapp} & \cellcolor{cotapp}4.73 & \cellcolor{cotapp}125 & \cellcolor{cotapp} & \cellcolor{cotapp}\textbf{3.49} & \cellcolor{cotapp}42 & \cellcolor{cotapp} & \cellcolor{cotapp}\textbf{4.02} \\
\bottomrule
\end{tabular}
\vspace{-4mm}
\end{table}

\noindent\textbf{The Repeat Curse extends to a third model and to long responses.} Table~\ref{tabrepeat} completes \S\ref{sec:moti_metric} with LaViDa and with $L{=}512$. Without a cache, no model repeats in more than $4.6\%$ of its responses at any length. Under dLLM-Cache, all three models repeat in more than $97\%$ of them at $L{=}512$, and a single run reaches $465$ tokens. On LLaDA-V and LaViDa, whose responses lengthen with the budget (Table~\ref{tabphrase}), the SRR rises with $L$, from $70.8\%$ and $24.6\%$ at $L{=}64$. MMaDA, whose responses level off near $100$ tokens, repeats at every length. On average over the lengths, SlowFast is markedly milder than dLLM-Cache on LLaDA-V alone.

\noindent\textbf{The damage lies in a few runs that consume the response.} Under dLLM-Cache the mean run is $2$--$6$ tokens and $95\%$ of the runs are no longer than $5$, so most runs are as short as those that occur without a cache. The longest run, however, grows with the budget, from $30$--$47$ tokens at $L{=}64$ to $409$--$465$ at $L{=}512$, that is, $80$--$91\%$ of the $512$-token budget; under SlowFast it reaches $232$--$428$ tokens at $L{=}512$. The Repeat Curse is not a matter of a few extra pairs but of responses that break into a single run.

\noindent\textbf{CoTA++ mitigates the repetition across models, backends and lengths.} Averaged over the three lengths and both backends, CoTA++ reduces the SRR of the cached model by $95\%$ on LLaDA-V, $88\%$ on LaViDa and $99\%$ on MMaDA, and its ARR by $99.5\%$, $97.7\%$ and $99.9\%$. No run longer than $3$ tokens remains in any setting. On average, it also repeats less than CoTA in every model-backend pair, and on LLaDA-V the margin widens as the responses lengthen: under dLLM-Cache, CoTA leaves an SRR of $2.4\%$ at $L{=}64$ and $47.4\%$ at $L{=}512$, where CoTA++ leaves $0.4\%$ and $8.2\%$, and under SlowFast CoTA still leaves a run of $286$ tokens at $L{=}512$.

\noindent\textbf{The repetition is not displaced to longer spans.} A remedy that acts on adjacent repeats could push the repetition into phrases. Table~\ref{tab:phrase512} shows that it does not: under dLLM-Cache at $L{=}512$, the share of repeated $4$-grams rises from $24.6\%$, $2.2\%$ and $2.0\%$ without a cache to $37.9\%$, $7.7\%$ and $14.5\%$ on LLaDA-V, MMaDA and LaViDa, and CoTA++ brings it back to $20.5\%$, $2.2\%$ and $1.9\%$, at or below the uncached value. Distinct-$2$ returns the same way.

\begin{table}[t]
\centering
\caption{\textbf{Phrase-level repetition} at $L{=}512$ under dLLM-Cache: repeated $4$-grams (Rep-4) and distinct $2$-grams (Dist-2) in percent (\S\ref{setDetails}). Bold: best among the cached rows. Shaded: CoTA++.}
\label{tab:phrase512}
\footnotesize
\setlength{\tabcolsep}{4pt}
\renewcommand{\arraystretch}{1.12}
\begin{tabular}{l cc cc cc}
\toprule
& \multicolumn{2}{c}{LLaDA-V} & \multicolumn{2}{c}{MMaDA} & \multicolumn{2}{c}{LaViDa} \\
\cmidrule(lr){2-3} \cmidrule(lr){4-5} \cmidrule(lr){6-7}
Method & Rep-4$\downarrow$ & Dist-2$\uparrow$ & Rep-4$\downarrow$ & Dist-2$\uparrow$ & Rep-4$\downarrow$ & Dist-2$\uparrow$ \\
\midrule
\textcolor{refgray}{Vanilla} & 24.6 & 58.2 & 2.2 & 89.3 & 2.0 & 86.8 \\
dLLM-Cache & 37.9 & 45.7 & 7.7 & 78.5 & 14.5 & 73.6 \\
\quad$+$\,CoTA & 21.9 & 57.7 & 2.4 & 88.2 & 2.0 & 86.6 \\
\cellcolor{cotapp}\quad$+$\,\textbf{CoTA++} & \cellcolor{cotapp}\textbf{20.5} & \cellcolor{cotapp}\textbf{59.1} & \cellcolor{cotapp}\textbf{2.2} & \cellcolor{cotapp}\textbf{89.2} & \cellcolor{cotapp}\textbf{1.9} & \cellcolor{cotapp}\textbf{86.9} \\
\bottomrule
\end{tabular}
\vspace{-4mm}
\end{table}

\noindent\textbf{The repetition is mitigated at no cost in quality.} Averaged over the three lengths, the cache lowers the judge score by $0.22$ to $1.23$ points (Table~\ref{tabphrase}). CoTA++ raises it in all six model-backend pairs, to within $0.34$ of the uncached model, and to within $0.04$ in four of them. On the same average, it scores at least as high as CoTA in every pair. The gains are not bought with shorter responses: on LLaDA-V under SlowFast, the one pair where the cache halves the response at $L{=}512$, CoTA++ answers at the length of the SlowFast baseline, $238$ against $254$ tokens, so its lower SRR is not an effect of length.

\begin{figure*}[!t]
\centering
\includegraphics[width=\textwidth]{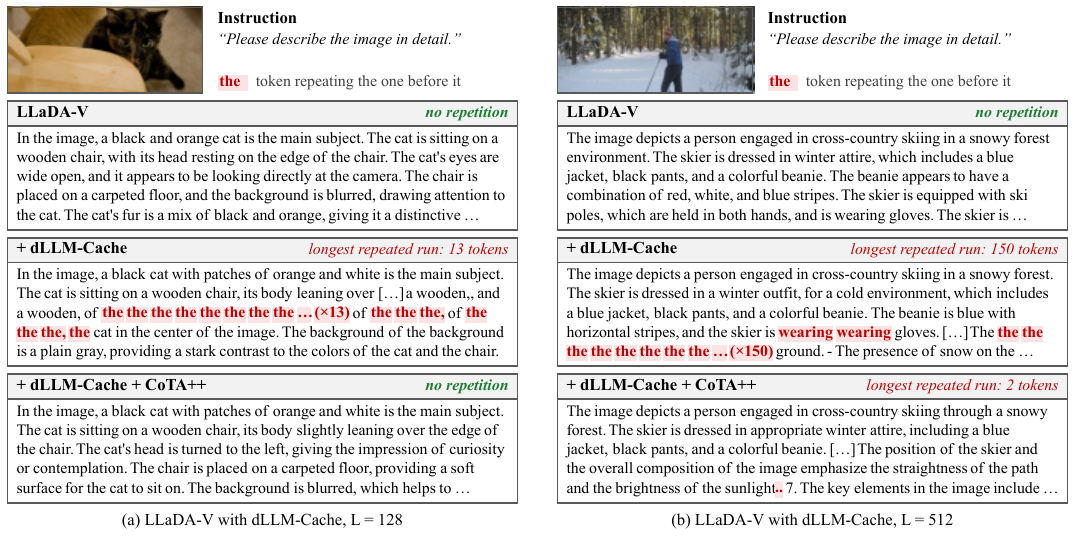}
\vspace{-5mm}
\caption{\textbf{Case study}, LLaDA-V under dLLM-Cache: the uncached model, the cached model, and the cached model with CoTA++. (a) The typical case, $L{=}128$. (b) A residual failure, $L{=}512$: a doubled period at a sentence boundary.}
\label{fig:case}
\end{figure*}

\noindent\textbf{Case study.} Fig.~\ref{fig:case} shows two responses of LLaDA-V under dLLM-Cache. In (a), at $L{=}128$, the cached model breaks into a run of $13$ identical tokens, and CoTA++ returns a response free of repetition. In (b), at $L{=}512$, the cache breaks into a run of $150$ copies of \emph{the}, and CoTA++ recovers the description except for a doubled period at a sentence boundary. This is the form of the curse that survives: where CoTA++ still repeats, no run is longer than $3$ tokens (Table~\ref{tabrepeat}), and the repeated tokens are punctuation, an article or correct text, such as \emph{a trash can can be seen} or the digits of a number in the image.

\begin{table*}[t]
\centering
\caption{\textbf{General multimodal benchmarks} under (a) dLLM-Cache and (b) SlowFast: DocVQA, ChartQA, MMStar, MME (perception, cognition), SEED, MMBench, LLaVA-Bench (in-the-wild), MathVista and MathVerse. MMaDA uses the MMaDA-8B-MixCoT checkpoint, the only one with a released evaluation protocol.}
\label{mllms}
\scriptsize
\setlength{\tabcolsep}{0.5pt}
\renewcommand{\arraystretch}{1.10}
\begin{minipage}[t]{0.49\textwidth}
\centering
\begin{tabular}{@{}l cccccccccc@{}}
\toprule
\multicolumn{11}{c}{(a) dLLM-Cache} \\
\midrule
Model & Doc & Chart & MMS & MME$^{\mathrm{P}}$ & MME$^{\mathrm{C}}$ & SEED & MMB & LLaVA$^{W}$ & MVista & MVerse \\
\midrule
LLaDA-V~\cite{lladav} & 83.9 & 78.3 & 60.2 & 1508 & 492 & 74.8 & 83.0 & 51.7 & 60.4 & 29.2 \\
\hspace{0.3em}$+$\,dLLM-Cache & 70.5 & 78.2 & 60.0 & 1503 & 497 & 74.8 & 83.0 & 30.2 & 55.9 & 26.1 \\
\hspace{0.3em}$+$\,CoTA~\cite{cota} & 81.6 & 78.6 & 60.0 & 1505 & 499 & 74.8 & 83.2 & 50.0 & 57.0 & 27.5 \\
\rowcolor{cotapp}
\hspace{0.3em}$+$\,\textbf{CoTA++} & 82.1 & 78.6 & 60.0 & 1505 & 499 & 74.8 & 83.2 & 50.4 & 57.2 & 28.3 \\
\midrule
MMaDA~\cite{mmada} & -- & 10.1 & 37.3 & 1087 & 285 & 56.6 & 42.6 & 29.7 & 34.2 & 18.8 \\
\hspace{0.3em}$+$\,dLLM-Cache & -- & 9.0 & 37.1 & 1082 & 274 & 56.7 & 42.1 & 33.2 & 34.5 & 20.4 \\
\hspace{0.3em}$+$\,CoTA~\cite{cota} & -- & 9.6 & 37.1 & 1086 & 277 & 56.7 & 42.5 & 33.9 & 34.5 & 20.8 \\
\rowcolor{cotapp}
\hspace{0.3em}$+$\,\textbf{CoTA++} & -- & 9.6 & 37.1 & 1086 & 277 & 56.7 & 42.5 & 35.0 & 35.3 & 21.4 \\
\midrule
LaViDa~\cite{lavida} & -- & 67.3 & 45.9 & 1346 & 368 & 66.4 & 71.7 & 50.5 & 44.9 & 22.1 \\
\hspace{0.3em}$+$\,dLLM-Cache & -- & 64.9 & 45.4 & 1339 & 334 & 66.2 & 71.4 & 45.8 & 44.6 & 15.8 \\
\hspace{0.3em}$+$\,CoTA~\cite{cota} & -- & 66.4 & 45.5 & 1345 & 338 & 66.1 & 71.6 & 45.9 & 45.2 & 18.8 \\
\rowcolor{cotapp}
\hspace{0.3em}$+$\,\textbf{CoTA++} & -- & 66.4 & 45.5 & 1345 & 338 & 66.1 & 71.6 & 47.1 & 45.6 & 19.6 \\
\bottomrule
\end{tabular}
\end{minipage}\hfill
\begin{minipage}[t]{0.49\textwidth}
\centering
\begin{tabular}{@{}l cccccccccc@{}}
\toprule
\multicolumn{11}{c}{(b) SlowFast} \\
\midrule
Model & Doc & Chart & MMS & MME$^{\mathrm{P}}$ & MME$^{\mathrm{C}}$ & SEED & MMB & LLaVA$^{W}$ & MVista & MVerse \\
\midrule
LLaDA-V~\cite{lladav} & 83.9 & 78.3 & 60.2 & 1508 & 492 & 74.8 & 83.0 & 51.7 & 60.4 & 29.2 \\
\hspace{0.3em}$+$\,SlowFast & 68.4 & 78.4 & 60.0 & 1503 & 497 & 74.8 & 83.0 & 37.6 & 54.2 & 26.7 \\
\hspace{0.3em}$+$\,CoTA~\cite{cota} & 79.1 & 78.6 & 60.2 & 1505 & 499 & 74.8 & 83.2 & 52.7 & 57.0 & 27.4 \\
\rowcolor{cotapp}
\hspace{0.3em}$+$\,\textbf{CoTA++} & 80.8 & 78.6 & 60.2 & 1505 & 499 & 74.8 & 83.2 & 53.5 & 57.2 & 28.0 \\
\midrule
MMaDA~\cite{mmada} & -- & 10.1 & 37.3 & 1087 & 285 & 56.6 & 42.6 & 29.7 & 34.2 & 18.8 \\
\hspace{0.3em}$+$\,SlowFast & -- & 9.2 & 37.1 & 1082 & 274 & 56.7 & 42.1 & 30.5 & 33.7 & 20.0 \\
\hspace{0.3em}$+$\,CoTA~\cite{cota} & -- & 9.7 & 37.2 & 1086 & 276 & 56.7 & 42.5 & 33.6 & 34.5 & 20.8 \\
\rowcolor{cotapp}
\hspace{0.3em}$+$\,\textbf{CoTA++} & -- & 9.7 & 37.2 & 1086 & 277 & 56.7 & 42.5 & 34.4 & 35.1 & 21.6 \\
\midrule
LaViDa~\cite{lavida} & -- & 67.3 & 45.9 & 1346 & 368 & 66.4 & 71.7 & 50.5 & 44.9 & 22.1 \\
\hspace{0.3em}$+$\,SlowFast & -- & 61.0 & 45.7 & 1339 & 345 & 66.1 & 71.9 & 41.4 & 44.0 & 15.1 \\
\hspace{0.3em}$+$\,CoTA~\cite{cota} & -- & 63.9 & 45.7 & 1345 & 359 & 66.1 & 72.2 & 46.1 & 44.9 & 17.7 \\
\rowcolor{cotapp}
\hspace{0.3em}$+$\,\textbf{CoTA++} & -- & 63.9 & 45.7 & 1345 & 359 & 66.1 & 72.2 & 46.8 & 45.3 & 18.9 \\
\bottomrule
\end{tabular}
\end{minipage}
\vspace{-1.5mm}
\end{table*}

\subsection{Performance on General Multimodal Benchmarks}\label{exp:general}

\noindent\textbf{Short answers are left as they are.} MMStar, MME, SEED-Bench and MMBench ask for an option letter or a single word, which leaves no room for repetition. On these benchmarks CoTA++ stays within $0.4$ points of the cached model, or $14$ points on the scale of MME, on all three models and under both backends (Table~\ref{mllms}). An intervention aimed at long responses thus costs nothing where there is nothing to repair.

\noindent\textbf{Longer answers are largely recovered.} The cache costs most where the answer is long. Under dLLM-Cache (Table~\ref{mllms}(a)), LLaDA-V loses $13.4$ points on DocVQA and $21.5$ on LLaVA-Bench, and CoTA++ returns both to within $1.8$ points of the uncached model. On MathVerse, CoTA++ recovers $2.2$ of the $3.1$ points lost on LLaDA-V and $3.8$ of the $6.3$ lost on LaViDa. Recovery is partial on MathVista, where CoTA++ returns $1.3$ of the $4.5$ points LLaDA-V loses, and on LLaVA-Bench for LaViDa, $1.3$ of $4.7$. On MMaDA the cache costs at most $1.1$ points on these benchmarks, and CoTA++ scores at or above the cached model on each. Under SlowFast (Table~\ref{mllms}(b)) the pattern is the same: on LLaDA-V, LLaVA-Bench returns from $37.6$ to $53.5$, above the uncached $51.7$, and DocVQA to within $3.1$ points; LaViDa recovers $2.9$ of the $6.3$ points on ChartQA, $5.4$ of $9.1$ on LLaVA-Bench and $3.8$ of $7.0$ on MathVerse; MMaDA scores at or above the backend on every benchmark.

\subsection{Component-wise Analysis of CoTA++}\label{exp:analysis}

\subsubsection{Ablation Study}\label{exp:ablation}

\begin{table}[t]
\centering
\caption{\textbf{Component grid} (LLaDA-V with dLLM-Cache, $500$ images): every subset of the three components at three response lengths. Shaded: CoTA++.}
\label{tab:component_grid}
\footnotesize
\setlength{\tabcolsep}{2pt}
\renewcommand{\arraystretch}{1.12}
\begin{tabular}{@{}ccc ccc ccc ccc@{}}
\toprule
 & & & \multicolumn{3}{c}{$L{=}64$} & \multicolumn{3}{c}{$L{=}128$} & \multicolumn{3}{c}{$L{=}512$} \\
\cmidrule(lr){4-6}\cmidrule(lr){7-9}\cmidrule(l){10-12}
{\scriptsize DAR} & {\scriptsize CTAR} & {\scriptsize CTEV} & ARR & SRR & MRL & ARR & SRR & MRL & ARR & SRR & MRL \\
\midrule
 &  &  & $1.61$ & $70.8$ & $38$ & $7.01$ & $96.8$ & $73$ & $14.70$ & $99.6$ & $455$ \\
\midrule
\cmark &  &  & $0.02$ & $1.4$ & $2$ & $0.04$ & $4.4$ & $2$ & $0.10$ & $11.2$ & $165$ \\
 & \cmark &  & $0.13$ & $8.0$ & $2$ & $0.21$ & $20.6$ & $3$ & $0.31$ & $62.0$ & $160$ \\
 &  & \cmark & $0.77$ & $35.6$ & $34$ & $0.86$ & $59.6$ & $56$ & $1.72$ & $94.0$ & $420$ \\
\midrule
\cmark & \cmark &  & $0.02$ & $0.6$ & $2$ & $0.02$ & $3.0$ & $2$ & $0.05$ & $9.6$ & $3$ \\
\cmark &  & \cmark & $0.09$ & $0.4$ & $2$ & $0.11$ & $6.6$ & $25$ & $0.03$ & $12.2$ & $3$ \\
 & \cmark & \cmark & $0.04$ & $2.4$ & $2$ & $0.10$ & $11.4$ & $2$ & $0.17$ & $47.4$ & $3$ \\
\cellcolor{gray!12}\cmark & \cellcolor{gray!12}\cmark & \cellcolor{gray!12}\cmark & \cellcolor{gray!12}$0.01$ & \cellcolor{gray!12}$0.4$ & \cellcolor{gray!12}$2$ & \cellcolor{gray!12}$0.02$ & \cellcolor{gray!12}$2.4$ & \cellcolor{gray!12}$2$ & \cellcolor{gray!12}$0.03$ & \cellcolor{gray!12}$8.2$ & \cellcolor{gray!12}$3$ \\
\midrule
\multicolumn{3}{@{}l}{Vanilla} & $0.01$ & $0.4$ & $2$ & $0.00$ & $0.2$ & $2$ & $0.01$ & $2.4$ & $3$ \\
\bottomrule
\end{tabular}
\end{table}

\begin{table}[t]
\centering
\caption{\textbf{Efficiency} on LLaDA-V at $L{=}64$, $20$ images, each configuration alone on one A100: latency in seconds per image, throughput in decoded tokens per second, latency relative to dLLM-Cache, and peak GPU memory.}
\label{tab:efficiency}
\footnotesize
\setlength{\tabcolsep}{4pt}
\renewcommand{\arraystretch}{1.12}
\begin{tabularx}{\columnwidth}{@{}l *{4}{>{\centering\arraybackslash}X}@{}}
\toprule
Configuration & \begin{tabular}[c]{@{}c@{}}Latency\\(s/img)$\downarrow$\end{tabular} & \begin{tabular}[c]{@{}c@{}}Throughput\\(tok/s)$\uparrow$\end{tabular} & \begin{tabular}[c]{@{}c@{}}Relative\\latency\end{tabular} & \begin{tabular}[c]{@{}c@{}}Memory\\(GiB)\end{tabular} \\
\midrule
\textcolor{refgray}{Vanilla} & 21.4 & 2.9 & -- & 30.6 \\
dLLM-Cache & 8.8 & 7.1 & 1.00$\times$ & 32.5 \\
\quad$+$\,DAR & 9.0 & 7.0 & 1.01$\times$ & 32.5 \\
\quad$+$\,CTAR & 9.4 & 6.7 & 1.06$\times$ & 32.5 \\
\quad$+$\,CTEV & 9.4 & 6.7 & 1.07$\times$ & 33.4 \\
\rowcolor{cotapp}\quad$+$\,\textbf{CoTA++} & 10.2 & 6.2 & 1.15$\times$ & 33.4 \\
\bottomrule
\end{tabularx}
\vspace{-1mm}
\end{table}
\begin{figure}[t]
\centering
\includegraphics[width=\columnwidth]{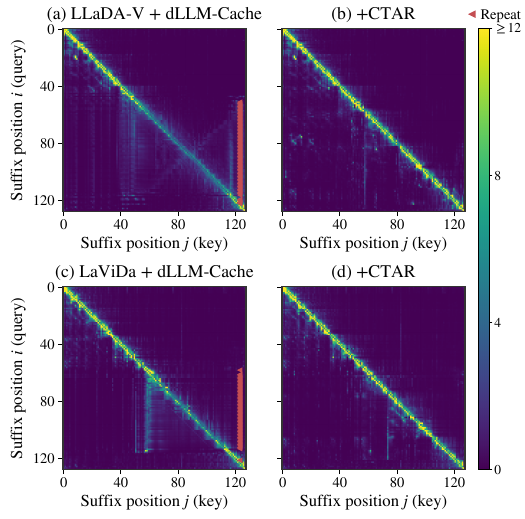}
\vspace{-4mm}
\caption{\textbf{Attention at the decode moment, with and without CTAR}, layers $25$--$32$; triangles: repeat positions. (a)--(b) LLaDA-V and (c)--(d) LaViDa under dLLM-Cache, $L{=}128$.}
\label{fig:attn_ctar}
\vspace{-4mm}
\end{figure}

\noindent\textbf{Each component alone mitigates the curse.} Table~\ref{tab:component_grid} evaluates every subset of the three components on LLaDA-V under dLLM-Cache. Alone, DAR is the strongest and CTEV the weakest: at $L{=}128$ they leave a repeat in $4.4\%$ and $59.6\%$ of the responses, CTAR in $20.6\%$, against $96.8\%$ under the cache. DAR and CTAR lower ARR by more than an order of magnitude at every length, CTEV by a factor of two to nine. None of them alone removes the tail at $L{=}512$, where a run of $165$, $160$ and $420$ tokens survives.

\noindent\textbf{Pairs remove the tail, and the full method is the most consistent across lengths.} Any two components remove the tail at $L{=}512$, where no pair leaves a run longer than $3$ tokens, although DAR with CTEV still leaves one of $25$ at $L{=}128$. The pairs differ in how many responses still repeat: DAR with CTAR leaves $9.6\%$ at $L{=}512$, CTAR with CTEV, the conference method, $47.4\%$. The full method is the strongest configuration at $L{=}128$ and $L{=}512$ and matches the best pair at $L{=}64$, DAR with CTEV ($0.4\%$): it leaves no run longer than $3$ tokens and an SRR of $0.4$, $2.4$ and $8.2\%$, against $0.6$, $3.0$ and $9.6\%$ for DAR with CTAR; the uncached model repeats in $0.4$, $0.2$ and $2.4\%$.

\subsubsection{Efficiency}\label{exp:efficiency}

\noindent\textbf{The components run at the speed of the backend they correct.} Table~\ref{tab:efficiency} times each configuration alone on one A100 at $L{=}64$. dLLM-Cache decodes $2.4$ times faster than the uncached model. On top of it, DAR is free ($1.01\times$), since it changes which positions the backend recomputes and not how many. CTAR costs $6\%$ for the rows its monitor re-forms in the eight deep layers, and CTEV $7\%$ for five logit-lens projections per step over the committed positions. Together they bring CoTA++ to $1.15\times$ the latency of the cache, with $0.9$\,GiB more memory, which is still $2.1$ times faster than decoding without a cache.

\subsubsection{CTAR: Routing at the Decode Moment}\label{exp:ctar}

\noindent F1 located the Repeat Curse in the routing of the within-step flow: under the cache, a position about to commit anchors less on its context tokens. Fig.~\ref{fig:attn_ctar} shows this at the decode moment. Without CTAR, the diagonal band of anchoring on the neighbors fades at the positions that repeat, and their attention spreads into vertical streaks on a few distant positions; with CTAR, the routing is re-formed against the current keys and the band returns as sharp as in the uncached model. The same loss and recovery appear on both models, so the dispersion belongs to cached decoding rather than to a model, and re-forming the routing restores the anchoring that the cache weakened.

\subsubsection{DAR: Timeliness of the KV State}\label{exp:dar}
\begin{figure}[t]
\centering
\includegraphics[width=\columnwidth]{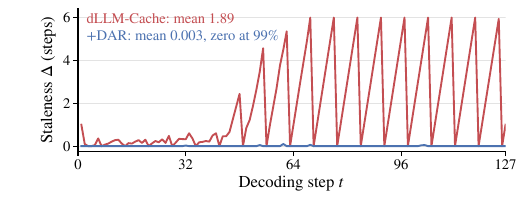}
\vspace{-4mm}
\caption{\textbf{Staleness of the KV state at the decode moment, with and without DAR.} Mean over $100$ images at each decoding step, dLLM-Cache, $L{=}128$.}
\label{fig:dar_stale}
\vspace{-2mm}
\end{figure}

\noindent Fig.~\ref{fig:dar_stale} reads the staleness of the KV state at the decode moment. Under the cache it follows the refresh interval, rising to six steps between refreshes; with DAR it is zero at $99\%$ of the decode moments.

\subsubsection{CTEV: Consolidation of the Context}\label{exp:ctev}
\begin{figure}[t]
\centering
\includegraphics[width=\columnwidth]{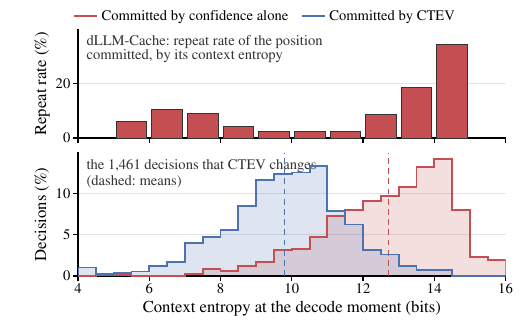}
\vspace{-4mm}
\caption{\textbf{Context entropy at the decode moment, with and without CTEV}, dLLM-Cache, $L{=}128$. Top: repeat rate of the committed position. Bottom: the decisions CTEV changes.}
\label{fig:ctev_dec}
\vspace{-2.5mm}

\end{figure}

\noindent F3 located the Repeat Curse in the consolidation of the cross-layer flow: under the cache, repetitions arise in unconsolidated contexts. Fig.~\ref{fig:ctev_dec} shows this at the decode moment. The repeat rate of the committed position stays low over most of the range of its context entropy and climbs steeply at the top of the range; at the decisions CTEV changes, the position that confidence alone would commit lies in that region, and the position CTEV commits instead lies where the repeat rate is low. CTEV thus moves the commitment out of the unconsolidated contexts in which F3 found the repetitions arise.

\section{Discussion}\label{sec:discussion}

\subsection{What Repeats}\label{disc:what}

\begin{table}[t]
\centering
\caption{\textbf{Design decisions of the three components}, LLaDA-V under dLLM-Cache, $L{=}128$, $100$ images, each component alone.}
\label{tab:design}
\footnotesize
\setlength{\tabcolsep}{3pt}
\renewcommand{\arraystretch}{1.12}
\begin{tabular}{l cccc}
\toprule
Configuration & ARR$\downarrow$ & SRR$\downarrow$ & MRL$\downarrow$ & Rep-4$\downarrow$ \\
\midrule
\textcolor{refgray}{Vanilla} & 0.00 & 0.0 & 0 & 0.037 \\
dLLM-Cache & 7.52 & 96.0 & 73 & 0.083 \\
\multicolumn{5}{@{}l}{\emph{CTAR: re-weight the stored routing, or re-form it}} \\
\quad Additive distance bias & 6.41 & 98.0 & 73 & 0.073 \\
\quad Gated gain & 9.07 & 99.0 & 82 & 0.086 \\
\rowcolor{cotapp}\quad Re-formed, layers $25$--$32$ (CTAR) & 0.32 & 21.0 & 3 & 0.035 \\
\multicolumn{5}{@{}l}{\emph{DAR: how many slots $r$, and what fills six slots}} \\
\quad $r{=}2$ & 0.39 & 36.0 & 2 & 0.022 \\
\rowcolor{cotapp}\quad $r{=}4$ (DAR) & 0.04 & 4.0 & 2 & 0.026 \\
\quad $r{=}16$ & 0.02 & 1.0 & 2 & 0.030 \\
\quad Dispersed anchoring (F1) & 7.66 & 98.0 & 72 & 0.083 \\
\quad Unconsolidated entropy (F3) & 1.31 & 57.0 & 77 & 0.038 \\
\quad Confidence (DAR) & 0.03 & 3.0 & 2 & 0.025 \\
\multicolumn{5}{@{}l}{\emph{CTEV: whose entropy, and how strongly}} \\
\quad Own entropy, $\lambda{=}0.25$ & 5.14 & 98.0 & 47 & 0.047 \\
\quad Context entropy, veto & 7.66 & 96.0 & 73 & 0.085 \\
\rowcolor{cotapp}\quad Context entropy, $\lambda{=}0.25$ (CTEV) & 0.71 & 61.0 & 3 & 0.031 \\
\quad Context entropy, $\lambda{=}0.5$ & 0.36 & 36.0 & 2 & 0.044 \\
\quad Context entropy, layers $1$--$10$ & 3.30 & 79.0 & 4 & 0.039 \\
\bottomrule
\end{tabular}
\vspace{-2mm}
\end{table}

\begin{figure}[t]
\centering
\includegraphics[width=\columnwidth]{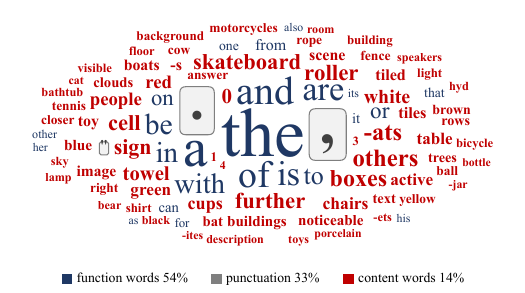}
\vspace{-4mm}
\caption{\textbf{Tokens that open a repeated run under the cache}, sized by run count; a leading hyphen marks a word piece.}
\label{fig:wordcloud}
\vspace{-2.5mm}

\end{figure}

\noindent Fig.~\ref{fig:wordcloud} collects the tokens that open a repeated run over the cached rows of Table~\ref{tabrepeat}. Function words open $54\%$ of the runs, punctuation $33\%$ and content words $14\%$: what repeats is the token whose identity is fixed by its context, the first to go wrong when the anchoring on that context weakens (F1).

\subsection{Design Alternatives}\label{disc:design}
\noindent\textbf{CTAR: re-forming the routing, not re-weighting it.} In Table~\ref{tab:design}, an additive distance bias leaves the curse in place and a gated gain raises ARR, whereas re-forming the routing against the current keys lowers ARR by more than an order of magnitude and removes the tail of long runs. The deep band follows F1; shallower bands trade a higher phrase-level statistic, and the full stack four times the rows, for a few more repeat-free responses (see the supplementary material).

\noindent\textbf{DAR: how many slots, and chosen by what.} Two reserved slots leave a third of the responses with a repeat, four bring SRR to $4\%$, and sixteen lower it further while the phrase-level statistic begins to rise (Table~\ref{tab:design}). The slots are worth reserving only for the positions about to commit: filled by the positions of most dispersed anchoring (F1), the cache is left as it was; by the least consolidated (F3), more than half of the responses still repeat; by confidence, $3\%$ do.

\noindent\textbf{CTEV: whose entropy, and how strongly.} Penalizing a candidate by its own entropy leaves the curse in place, and a veto on the context entropy reproduces the cache (Table~\ref{tab:design}). The graded penalty mitigates the curse, and $\lambda$ trades the two statistics: $0.5$ lowers the share of responses with a repeat from $61\%$ to $36\%$ but raises the phrase-level statistic above the uncached level. Read from the shallow layers instead of the deep window, the same penalty leaves ARR above $3$.

\noindent\textbf{Fixed budgets and adaptive budgets.} A single token-level risk score could instead merge the three signals and allocate one compute budget adaptively. The three components already act per position, each steered by a signal of its own disruption, the monitor count of CTAR, the confidence rank of DAR and the context entropy of CTEV, and only the amount spent per step, set by $\kappa$, $r$ and $\lambda$, is fixed. Such a score presumes that the signals are interchangeable, which Table~\ref{tab:design} rules out: DAR slots filled by the F1 or F3 signal leave the curse largely in place, and a veto on the F3 signal reproduces the cache, because the disruptions act on different objects. A risk score could still set how much of each budget to spend at a step, for instance more reserved slots when more positions fall in the top fifth by context entropy, which we leave to future work.

\subsection{Decoding-Time Alternatives and Settings}\label{disc:settings}
\noindent\textbf{A decoding-time penalty caps the runs and leaves the curse.} Applying the no-repeat $n$-gram rule of autoregressive decoding~\cite{paulus2017deep} to LLaDA-V under dLLM-Cache, which blocks any token that would complete a $2$-gram or $3$-gram already in the response, caps the longest run at a few tokens, yet nearly all responses still carry a repeat and the judge score does not improve, whereas CoTA++ leaves a repeat in a few percent of them at a higher score (see the supplementary material).

\noindent\textbf{Block-wise decoding leaves the picture unchanged.} In blocks of $16$ to $128$ positions, the cache gets worse the shorter the block, and CoTA++ returns every setting close to the uncached model (see the supplementary material).

\section{Conclusion}\label{sec:conclusion}
This work traces the Repeat Curse of cached dMLLMs to three disruptions of the context-token information flow: a position about to commit anchors less on its context tokens (F1), is decoded on a stale KV state (F2), and commits from an unconsolidated context (F3). CoTA++ counteracts each where it originates, without training, and across three dMLLMs, two caching backends and three response lengths it mitigates the curse at no cost in quality and keeps the speedup of caching. The broader lesson is that an approximate cache must preserve the information flow it accelerates, not only the freshness of its states. A refresh schedule chosen by feature similarity keeps the states fresh on average, yet lets the flow break at the positions about to be committed, and the repair has to follow the flow: within the step, across steps and across layers.

Open-weight dMLLMs and caching methods designed for them are still few, and our compute is limited, so CoTA++ has not yet been evaluated on further caching methods or larger dMLLMs. These are the first steps of future work. Broader ones are to allocate the three budgets adaptively from their signals (\S\ref{disc:design}) and to let the three findings inform how dMLLMs are trained and how their caches are designed.

\section*{Acknowledgment}
Generative AI tools, including Grammarly and Claude, were used solely for grammar checking and language polishing. The originality of this work was not influenced by generative AI. All core designs, data analysis, experimental findings, and technical contributions were independently developed by the human authors.

\bibliographystyle{IEEEtran}
\bibliography{ref}

@inproceedings{cota,
title={Context Tokens are Anchors: Understanding the Repeat Curse in d{MLLM}s from an Information Flow Perspective},
author={Qiyan Zhao and Xiaofeng Zhang and Shuochen Chang and Qianyu Chen and Xiaosong Yuan and Xuhang Chen and Luoqi Liu and Jiajun Zhang and Xu-Yao Zhang and Da-Han Wang},
booktitle={The Fourteenth International Conference on Learning Representations},
year={2026},
}

@inproceedings{llada,
title={Large Language Diffusion Models},
author={Shen Nie and Fengqi Zhu and Zebin You and Xiaolu Zhang and Jingyang Ou and Jun Hu and JUN ZHOU and Yankai Lin and Ji-Rong Wen and Chongxuan Li},
booktitle={The Thirty-ninth Annual Conference on Neural Information Processing Systems},
year={2025},
}

@misc{llada1.5,
      title={LLaDA 1.5: Variance-Reduced Preference Optimization for Large Language Diffusion Models}, 
      author={Fengqi Zhu and Rongzhen Wang and Shen Nie and Xiaolu Zhang and Chunwei Wu and Jun Hu and Jun Zhou and Jianfei Chen and Yankai Lin and Ji-Rong Wen and Chongxuan Li},
      year={2025},
      eprint={2505.19223},
      archivePrefix={arXiv},
      primaryClass={cs.LG},
}

@misc{dream2025,
    title = {Dream 7B},
    url = {https://hkunlp.github.io/blog/2025/dream},
    author = {Ye, Jiacheng and Xie, Zhihui and Zheng, Lin and Gao, Jiahui and Wu, Zirui and Jiang, Xin and Li, Zhenguo and Kong, Lingpeng},
    year = {2025}
}

@article{bie2025llada2,
  title={Llada2. 0: Scaling up diffusion language models to 100b},
  author={Bie, Tiwei and Cao, Maosong and Chen, Kun and Du, Lun and Gong, Mingliang and Gong, Zhuochen and Gu, Yanmei and Hu, Jiaqi and Huang, Zenan and Lan, Zhenzhong and others},
  journal={arXiv preprint arXiv:2512.15745},
  year={2025}
}

@article{llama3,
  title={The llama 3 herd of models},
  author={Grattafiori, Aaron and Dubey, Abhimanyu and Jauhri, Abhinav and Pandey, Abhinav and Kadian, Abhishek and Al-Dahle, Ahmad and Letman, Aiesha and Mathur, Akhil and Schelten, Alan and Vaughan, Alex and others},
  journal={arXiv preprint arXiv:2407.21783},
  year={2024}
}

@article{qwen2.5,
  author={An Yang and Baosong Yang and Beichen Zhang and Binyuan Hui and Bo Zheng and Bowen Yu and Chengyuan Li and others},
  title={Qwen2.5 Technical Report},
  year={2024},
  journal={CoRR},
  volume={abs/2412.15115},
}

@article{lladav,
  title={LLaDA-V: Large Language Diffusion Models with Visual Instruction Tuning},
  author={You, Zebin and Nie, Shen and Zhang, Xiaolu and Hu, Jun and Zhou, Jun and Lu, Zhiwu and Wen, Ji-Rong and Li, Chongxuan},
  journal={arXiv preprint arXiv:2505.16933},
  year={2025}
}

@inproceedings{
mmada,
title={{MM}a{DA}: Multimodal Large Diffusion Language Models},
author={Ling Yang and Ye Tian and Bowen Li and Xinchen Zhang and Ke Shen and Yunhai Tong and Mengdi Wang},
booktitle={The Thirty-ninth Annual Conference on Neural Information Processing Systems},
year={2025},
}

@inproceedings{
lavida,
title={LaViDa: A Large Diffusion Model for Vision-Language Understanding},
author={Shufan Li and Konstantinos Kallidromitis and Hritik Bansal and Akash Gokul and Yusuke Kato and Kazuki Kozuka and Jason Kuen and Zhe Lin and Kai-Wei Chang and Aditya Grover},
booktitle={The Thirty-ninth Annual Conference on Neural Information Processing Systems},
year={2025},
}

@inproceedings{li2026lavida,
  title={Lavida-o: Elastic large masked diffusion models for unified multimodal understanding and generation},
  author={Li, Shufan and Gu, Jiuxiang and Liu, Kangning and Lin, Zhe and Wei, Zijun and Grover, Aditya and Kuen, Jason},
  booktitle={International Conference on Learning Representations},
  volume={2026},
  pages={50571--50603},
  year={2026}
}

@article{zeng2025diffusionvl,
  title={Diffusionvl: Translating any autoregressive models into diffusion vision language models},
  author={Zeng, Lunbin and Yao, Jingfeng and Liao, Bencheng and Tao, Hongyuan and Liu, Wenyu and Wang, Xinggang},
  journal={arXiv preprint arXiv:2512.15713},
  year={2025}
}

@article{ai2026llada2,
  title={Llada2. 0-uni: Unifying multimodal understanding and generation with diffusion large language model},
  author={AI, Inclusion and Bie, Tiwei and Chen, Haoxing and Chen, Tieyuan and Cheng, Zhenglin and Cui, Long and Gan, Kai and Huang, Zhicheng and Lan, Zhenzhong and Li, Haoquan and others},
  journal={arXiv preprint arXiv:2604.20796},
  year={2026}
}

@article{xin2025lumina,
  title={Lumina-dimoo: An omni diffusion large language model for multi-modal generation and understanding},
  author={Xin, Yi and Qin, Qi and Luo, Siqi and Zhu, Kaiwen and Yan, Juncheng and Tai, Yan and Lei, Jiayi and Cao, Yuewen and Wang, Keqi and Wang, Yibin and others},
  journal={arXiv preprint arXiv:2510.06308},
  year={2025}
}

@article{llava1.5,
  title={Improved Baselines with Visual Instruction Tuning},
  author={Haotian Liu and Chunyuan Li and Yuheng Li and Yong Jae Lee},
  journal={2024 IEEE/CVF Conference on Computer Vision and Pattern Recognition (CVPR)},
  year={2023},
  pages={26286-26296},
}

@inproceedings{
dLLM-Cache,
title={d{LLM}-Cache: Accelerating Diffusion Large Language Models with Adaptive Caching},
author={Zhiyuan Liu and Yicun Yang and Yaojie Zhang and Junjie Chen and Chang Zou and Qingyan Wei and Shaobo Wang and Yichen Zhu and Linfeng Zhang},
booktitle={Forty-third International Conference on Machine Learning},
year={2026},
}

@inproceedings{xiao2024streamingllm,
  title={Efficient streaming language models with attention sinks},
  author={Xiao, Guangxuan and Tian, Yuandong and Chen, Beidi and Han, Song and Lewis, Mike},
  booktitle={International Conference on Learning Representations},
  volume={2024},
  pages={21875--21895},
  year={2024}
}

@inproceedings{chang2026d3tom,
  title={D3tom: Decider-guided dynamic token merging for accelerating diffusion mllms},
  author={Chang, Shuochen and Zhang, Xiaofeng and Liu, Qingyang and Niu, Li},
  booktitle={Proceedings of the AAAI Conference on Artificial Intelligence},
  volume={40},
  number={24},
  pages={19961--19969},
  year={2026}
}

@inproceedings{xu2026fixing,
  title={Fixing Semantic Blind Spots in Anchor Tokens of dMLLMs},
  author={Xu, Ruixuan and Xu, Jiexi and Zhao, Qiyan and Zhang, Xiaofeng},
  booktitle={Findings of the Association for Computational Linguistics: ACL 2026},
  pages={19359--19374},
  year={2026}
}

@inproceedings{SlowFast,
  title={Accelerating diffusion large language models with slowfast sampling: The three golden principles},
  author={Wei, Qingyan and Zhang, Yaojie and Liu, Zhiyuan and Zeng, Puyu and Wang, Yuxuan and Qi, Biqing and Liu, Dongrui and Zhang, Linfeng},
  booktitle={International Conference on Learning Representations},
  volume={2026},
  pages={4340--4354},
  year={2026}
}

@inproceedings{Fast-dLLM,
  title={Fast-dllm: Training-free acceleration of diffusion llm by enabling kv cache and parallel decoding},
  author={Wu, Chengyue and Zhang, Hao and Xue, Shuchen and Liu, Zhijian and Diao, Shizhe and Zhu, Ligeng and Luo, Ping and Han, Song and Xie, Enze},
  booktitle={International Conference on Learning Representations},
  volume={2026},
  pages={57027--57051},
  year={2026}
}

@inproceedings{Lin2014MicrosoftCC,
  title={Microsoft COCO: Common Objects in Context},
  author={Tsung-Yi Lin and Michael Maire and Serge J. Belongie and James Hays and Pietro Perona and Deva Ramanan and Piotr Doll{\'a}r and C. Lawrence Zitnick},
  booktitle={European Conference on Computer Vision},
  year={2014},
}

@inproceedings{
Yu2024UnveilingAH,
title={Unveiling and Harnessing Hidden Attention Sinks: Enhancing Large Language Models without Training through Attention Calibration},
author={Zhongzhi Yu and Zheng Wang and Yonggan Fu and Huihong Shi and Khalid Shaikh and Yingyan Celine Lin},
booktitle={Forty-first International Conference on Machine Learning},
year={2024},
}

@inproceedings{Wang2023LabelWA,
    title = "Label Words are Anchors: An Information Flow Perspective for Understanding In-Context Learning",
    author = "Wang, Lean and Li, Lei  and Dai, Damai  and Chen, Deli  and Zhou, Hao  and Meng, Fandong  and Zhou, Jie  and Sun, Xu",
    booktitle = "Proceedings of the 2023 Conference on Empirical Methods in Natural Language Processing",
    year = "2023",
    pages = "9840--9855",
}

@inproceedings{fastv,
  title={An Image is Worth 1/2 Tokens After Layer 2: Plug-and-Play Inference Acceleration for Large Vision-Language Models},
  author={Liang Chen and Haozhe Zhao and Tianyu Liu and Shuai Bai and Junyang Lin and Chang Zhou and Baobao Chang},
  booktitle={European Conference on Computer Vision},
  year={2024},
}

@inproceedings{
Yona2025InterpretingTR,
title={Interpreting the Repeated Token Phenomenon in Large Language Models},
author={Itay Yona and Ilia Shumailov and Jamie Hayes and Yossi Gandelsman},
booktitle={Forty-second International Conference on Machine Learning},
year={2025},
}

@inproceedings{zhang2025shallow,
  title={Shallow focus, deep fixes: Enhancing shallow layers vision attention sinks to alleviate hallucination in lvlms},
  author={Zhang, Xiaofeng and Quan, Yihao and Shen, Chen and Gu, Chaochen and Yuan, Xiaosong and Yan, Shaotian and Cao, Jiawei and Cheng, Hao and Wu, Kaijie and Ye, Jieping},
  booktitle={Proceedings of the 2025 Conference on Empirical Methods in Natural Language Processing},
  pages={3512--3534},
  year={2025}
}

@inproceedings{SDXL,
  title={Sdxl: Improving latent diffusion models for high-resolution image synthesis},
  author={Podell, Dustin and English, Zion and Lacey, Kyle and Blattmann, Andreas and Dockhorn, Tim and M{\"u}ller, Jonas and Penna, Joe and Rombach, Robin},
  booktitle={International Conference on Learning Representations},
  volume={2024},
  pages={1862--1874},
  year={2024}
}

@inproceedings{esser2024scaling,
  title={Scaling rectified flow transformers for high-resolution image synthesis},
  author={Esser, Patrick and Kulal, Sumith and Blattmann, Andreas and Entezari, Rahim and M{\"u}ller, Jonas and Saini, Harry and Levi, Yam and Lorenz, Dominik and Sauer, Axel and Boesel, Frederic and others},
  booktitle={Forty-first international conference on machine learning},
  year={2024}
}

@inproceedings{
zhang2025target,
title={Target Concrete Score Matching: A Holistic Framework for Discrete Diffusion},
author={Ruixiang ZHANG and Shuangfei Zhai and Yizhe Zhang and James Thornton and Zijing Ou and Joshua M. Susskind and Navdeep Jaitly},
booktitle={Forty-second International Conference on Machine Learning},
year={2025},
}

@inproceedings{
gong2023diffuseq,
title={DiffuSeq: Sequence to Sequence Text Generation with Diffusion Models},
author={Shansan Gong and Mukai Li and Jiangtao Feng and Zhiyong Wu and Lingpeng Kong},
booktitle={The Eleventh International Conference on Learning Representations },
year={2023},
}

@inproceedings{zheng2025masked,
  title={Masked diffusion models are secretly time-agnostic masked models and exploit inaccurate categorical sampling},
  author={Zheng, Kaiwen and Chen, Yongxin and Mao, Hanzi and Liu, Ming-Yu and Zhu, Jun and Zhang, Qinsheng},
  booktitle={International Conference on Learning Representations},
  volume={2025},
  pages={63186--63227},
  year={2025}
}

@inproceedings{ou2025your,
  title={Your absorbing discrete diffusion secretly models the conditional distributions of clean data},
  author={Ou, Jingyang and Nie, Shen and Xue, Kaiwen and Zhu, Fengqi and Sun, Jiacheng and Li, Zhenguo and Li, Chongxuan},
  booktitle={International Conference on Learning Representations},
  volume={2025},
  pages={64972--65009},
  year={2025}
}

@inproceedings{nie2025scaling,
  title={Scaling up masked diffusion models on text},
  author={Nie, Shen and Zhu, Fengqi and Du, Chao and Pang, Tianyu and Liu, Qian and Zeng, Guangtao and Lin, Min and Li, Chongxuan},
  booktitle={International Conference on Learning Representations},
  volume={2025},
  pages={82974--82997},
  year={2025}
}

@article{xie2025dream,
  title={Dream-coder 7b: An open diffusion language model for code},
  author={Xie, Zhihui and Ye, Jiacheng and Zheng, Lin and Gao, Jiahui and Dong, Jingwei and Wu, Zirui and Zhao, Xueliang and Gong, Shansan and Jiang, Xin and Li, Zhenguo and others},
  journal={arXiv preprint arXiv:2509.01142},
  year={2025}
}

@article{zhu2025llada,
  title={LLaDA-MoE: A sparse MoE diffusion language model},
  author={Zhu, Fengqi and You, Zebin and Xing, Yipeng and Huang, Zenan and Liu, Lin and Zhuang, Yihong and Lu, Guoshan and Wang, Kangyu and Wang, Xudong and Wei, Lanning and others},
  journal={arXiv preprint arXiv:2509.24389},
  year={2025}
}

@article{Tschannen2025SigLIP2M,
  title={Siglip 2: Multilingual vision-language encoders with improved semantic understanding, localization, and dense features},
  author={Tschannen, Michael and Gritsenko, Alexey and Wang, Xiao and Naeem, Muhammad Ferjad and Alabdulmohsin, Ibrahim and Parthasarathy, Nikhil and Evans, Talfan and Beyer, Lucas and Xia, Ye and Mustafa, Basil and others},
  journal={arXiv preprint arXiv:2502.14786},
  year={2025}
}

@misc{Qwen2.5-VL,
      title={Qwen2.5-VL Technical Report}, 
      author={Shuai Bai and Keqin Chen and Xuejing Liu and Jialin Wang and others},
      year={2025},
      eprint={2502.13923},
      archivePrefix={arXiv},
      primaryClass={cs.CV},
}

@inproceedings{MLSYS2023_c4be71ab,
 author = {Pope, Reiner and Douglas, Sholto and Chowdhery, Aakanksha and Devlin, Jacob and Bradbury, James and Heek, Jonathan and Xiao, Kefan and Agrawal, Shivani and Dean, Jeff},
 booktitle = {Proceedings of Machine Learning and Systems},
 pages = {606--624},
 title = {Efficiently Scaling Transformer Inference},
 volume = {5},
 year = {2023}
}

@inproceedings{MaskKV,
  title={Mask tokens as prophet: Fine-grained cache eviction for efficient dllm inference},
  author={Huang, Jianuo and Zhang, Yaojie and Yang, Yicun and Huang, Benhao and Zhang, Linfeng},
  booktitle={Findings of the Association for Computational Linguistics: ACL 2026},
  pages={3456--3479},
  year={2026}
}

@article{dkvcache,
  title={dkv-cache: The cache for diffusion language models},
  author={Ma, Xinyin and Yu, Runpeng and Fang, Gongfan and Wang, Xinchao},
  journal={Advances in Neural Information Processing Systems},
  volume={38},
  pages={149009--149033},
  year={2026}
}

@inproceedings{FlashDLM,
  title={Flashdlm: Accelerating diffusion language model inference via efficient kv caching and guided diffusion},
  author={Hu, Zhanqiu and Meng, Jian and Akhauri, Yash and Abdelfattah, Mohamed and Seo, Jae-sun and Zhang, Zhiru and Gupta, Udit},
  booktitle={International Conference on Learning Representations},
  volume={2026},
  pages={153529--153544},
  year={2026}
}

@article{SeedDiffusion,
  title={Seed diffusion: A large-scale diffusion language model with high-speed inference},
  author={Song, Yuxuan and Zhang, Zheng and Luo, Cheng and Gao, Pengyang and Xia, Fan and Luo, Hao and Li, Zheng and Yang, Yuehang and Yu, Hongli and Qu, Xingwei and others},
  journal={arXiv preprint arXiv:2508.02193},
  year={2025}
}

@inproceedings{TimeFeature,
  title={Time is a feature: Exploiting temporal dynamics in diffusion language models},
  author={Wang, Wen and Fang, Bozhen and Jing, Chenchen and Shen, Yongliang and Shen, Yangyi and Wang, Qiuyu and Ouyang, Hao and Chen, Hao and Shen, Chunhua},
  booktitle={International Conference on Learning Representations},
  volume={2026},
  pages={72909--72935},
  year={2026}
}

@inproceedings{abnar2020quantifying,
  title={Quantifying attention flow in transformers},
  author={Abnar, Samira and Zuidema, Willem},
  booktitle={Proceedings of the 58th annual meeting of the association for computational linguistics},
  pages={4190--4197},
  year={2020}
}

@inproceedings{ferrando2024information,
  title={Information flow routes: Automatically interpreting language models at scale},
  author={Ferrando, Javier and Voita, Elena},
  booktitle={Proceedings of the 2024 Conference on Empirical Methods in Natural Language Processing},
  pages={17432--17445},
  year={2024}
}

@inproceedings{zhang2025cross,
  title={Cross-modal information flow in multimodal large language models},
  author={Zhang, Zhi and Yadav, Srishti and Han, Fengze and Shutova, Ekaterina},
  booktitle={2025 IEEE/CVF Conference on Computer Vision and Pattern Recognition (CVPR)},
  pages={19781--19791},
  year={2025},
  organization={IEEE}
}

@article{olsson2022context,
  title={In-context learning and induction heads},
  author={Olsson, Catherine and Elhage, Nelson and Nanda, Neel and Joseph, Nicholas and DasSarma, Nova and Henighan, Tom and Mann, Ben and Askell, Amanda and Bai, Yuntao and Chen, Anna and others},
  journal={arXiv preprint arXiv:2209.11895},
  year={2022}
}

@article{zhang2026fragile,
  title={The fragile truth of saliency: Improving LLM input attribution via attention bias optimization},
  author={Zhang, Yihua and Wang, Changsheng and Chen, Yiwei and Fan, Chongyu and Jia, Jinghan and Liu, Sijia},
  journal={Advances in Neural Information Processing Systems},
  volume={38},
  pages={154395--154421},
  year={2026}
}

@article{OPERA,
  title={OPERA: Alleviating Hallucination in Multi-Modal Large Language Models via Over-Trust Penalty and Retrospection-Allocation},
  author={Qidong Huang and Xiao-wen Dong and Pan Zhang and Bin Wang and Conghui He and Jiaqi Wang and Dahua Lin and Weiming Zhang and Neng H. Yu},
  journal={2024 IEEE/CVF Conference on Computer Vision and Pattern Recognition (CVPR)},
  year={2023},
  pages={13418-13427},
}

@inproceedings{zhaoMCA,
  title={Mca-llava: Manhattan causal attention for reducing hallucination in large vision-language models},
  author={Zhao, Qiyan and Zhang, Xiaofeng and Li, Yiheng and Xing, Yun and Yuan, Xiaosong and Tang, Feilong and Fan, Sinan and Chen, Xuhang and Wang, Da-Han and Zhang, Xu-Yao},
  booktitle={Proceedings of the 33rd ACM International Conference on Multimedia},
  pages={3981--3990},
  year={2025}
}

@article{qiu2026gated,
  title={Gated attention for large language models: Non-linearity, sparsity, and attention-sink-free},
  author={Qiu, Zihan and Wang, Zekun and Zheng, Bo and Huang, Zeyu and Wen, Kaiyue and Yang, Songlin and Men, Rui and Yu, Le and Huang, Fei and Huang, Suozhi and others},
  journal={Advances in Neural Information Processing Systems},
  volume={38},
  pages={100092--100118},
  year={2026}
}

@inproceedings{song2026sparse,
  title={Sparse-dllm: Accelerating diffusion llms with dynamic cache eviction},
  author={Song, Yuerong and Liu, Xiaoran and Li, Ruixiao and Liu, Zhigeng and Huang, Zengfeng and Guo, Qipeng and He, Ziwei and Qiu, Xipeng},
  booktitle={Proceedings of the AAAI Conference on Artificial Intelligence},
  volume={40},
  number={39},
  pages={33038--33046},
  year={2026}
}

@inproceedings{DPad,
  title={Dpad: Efficient diffusion language models with suffix dropout},
  author={Chen, Xinhua and Huang, Sitao and Guo, Cong and Wei, Chiyue and He, Yintao and Zhang, Jianyi and Li, Hai and Chen, Yiran},
  booktitle={International Conference on Learning Representations},
  volume={2026},
  pages={57286--57316},
  year={2026}
}

@inproceedings{zhang2025redundancy,
  title={From redundancy to relevance: Information flow in lvlms across reasoning tasks},
  author={Zhang, Xiaofeng and Quan, Yihao and Shen, Chen and Yuan, Xiaosong and Yan, Shaotian and Xie, Liang and Wang, Wenxiao and Gu, Chaochen and Tang, Hao and Ye, Jieping},
  booktitle={Proceedings of the 2025 Conference of the Nations of the Americas Chapter of the Association for Computational Linguistics: Human Language Technologies (Volume 1: Long Papers)},
  pages={2289--2299},
  year={2025}
}

@inproceedings{li2025token,
  title={Token activation map to visually explain multimodal llms},
  author={Li, Yi and Wang, Hualiang and Ding, Xinpeng and Wang, Haonan and Li, Xiaomeng},
  booktitle={2025 IEEE/CVF International Conference on Computer Vision (ICCV)},
  pages={48--58},
  year={2025},
  organization={IEEE}
}

@inproceedings{Yao2025UnderstandingTR,
  title={Understanding the repeat curse in large language models from a feature perspective},
  author={Yao, Junchi and Yang, Shu and Xu, Jianhua and Hu, Lijie and Li, Mengdi and Wang, Di},
  booktitle={Findings of the Association for Computational Linguistics: ACL 2025},
  pages={7787--7815},
  year={2025}
}

@inproceedings{nguyen2025beyond,
  title={Beyond semantic entropy: Boosting LLM uncertainty quantification with pairwise semantic similarity},
  author={Nguyen, Dang and Payani, Ali and Mirzasoleiman, Baharan},
  booktitle={Findings of the Association for Computational Linguistics: ACL 2025},
  pages={4530--4540},
  year={2025}
}

@article{gan2026unleashing,
  title={Unleashing diffusion transformers for visual correspondence by modulating massive activations},
  author={Gan, Chaofan and Tu, Yuanpeng and Chen, Xi and Chen, Tieyuan and Li, Yuxi and Harandi, Mehrtash and Lin, Weiyao},
  journal={Advances in Neural Information Processing Systems},
  volume={38},
  pages={114432--114462},
  year={2026}
}

@inproceedings{gan2026massive,
  title={Massive activations are the key to local detail synthesis in diffusion transformers},
  author={Gan, Chaofan and Zhao, Zicheng and Tu, Yuanpeng and Chen, Xi and Qin, Ziran and Chen, Tieyuan and Harandi, Mehrtash and Lin, Weiyao},
  booktitle={International Conference on Learning Representations},
  volume={2026},
  pages={104855--104883},
  year={2026}
}

@inproceedings{rulli2026attention,
  title={Attention sinks in diffusion language models},
  author={Rulli, Maximo Eduardo and Petruzzi, Simone and Michielon, Edoardo and Silvestri, Fabrizio and Scardapane, Simone and Devoto, Alessio},
  booktitle={Findings of the Association for Computational Linguistics: ACL 2026},
  pages={43244--43256},
  year={2026}
}

@inproceedings{zhang2026one,
  title={One token is enough: Improving diffusion language models with a sink token},
  author={Zhang, Zihou and Xie, Zheyong and Zhong, Li and Liu, Haifeng and Cao, Shaosheng},
  booktitle={Findings of the Association for Computational Linguistics: ACL 2026},
  pages={6479--6490},
  year={2026}
}

@inproceedings{zuo2026diffusion,
  title={Diffusion-CAM: Faithful Visual Explanations for dMLLMs},
  author={Zuo, Haomin and Li, Yidi and Yang, Luoxiao and Zhang, Xiaofeng},
  booktitle={Proceedings of the 64th Annual Meeting of the Association for Computational Linguistics (Volume 1: Long Papers)},
  pages={12085--12101},
  year={2026}
}

@inproceedings{hong2026mitigating,
title={Mitigating Mask Prior Drift and Positional Attention Collapse in Large Diffusion Vision-Language Models},
author={Sujung Hong and Chanyong Yoon and Seong Jae Hwang},
booktitle={Forty-third International Conference on Machine Learning},
year={2026},
}

@inproceedings{zhu2023penalty,
  title={Penalty decoding: Well suppress the self-reinforcement effect in open-ended text generation},
  author={Zhu, Wenhong and Hao, Hongkun and Wang, Rui},
  booktitle={Proceedings of the 2023 Conference on Empirical Methods in Natural Language Processing},
  pages={1218--1228},
  year={2023}
}

@article{su2022contrastive,
  title={A contrastive framework for neural text generation},
  author={Su, Yixuan and Lan, Tian and Wang, Yan and Yogatama, Dani and Kong, Lingpeng and Collier, Nigel},
  journal={Advances in neural information processing systems},
  volume={35},
  pages={21548--21561},
  year={2022}
}

@inproceedings{fu2021theoretical,
  title={A theoretical analysis of the repetition problem in text generation},
  author={Fu, Zihao and Lam, Wai and So, Anthony Man-Cho and Shi, Bei},
  booktitle={Proceedings of the AAAI Conference on Artificial Intelligence},
  volume={35},
  number={14},
  pages={12848--12856},
  year={2021}
}

@inproceedings{lagutinmplicit,
  title={Implicit unlikelihood training: Improving neural text generation with reinforcement learning},
  author={Lagutin, Evgeny and Gavrilov, Daniil and Kalaidin, Pavel},
  booktitle={Proceedings of the 16th Conference of the European Chapter of the Association for Computational Linguistics: Main Volume},
  pages={1432--1441},
  year={2021}
}

@article{DITTO,
  title={Learning to break the loop: Analyzing and mitigating repetitions for neural text generation},
  author={Xu, Jin and Liu, Xiaojiang and Yan, Jianhao and Cai, Deng and Li, Huayang and Li, Jian},
  journal={Advances in Neural Information Processing Systems},
  volume={35},
  pages={3082--3095},
  year={2022}
}

@article{li2023repetition,
  title={Repetition in repetition out: Towards understanding neural text degeneration from the data perspective},
  author={Li, Huayang and Lan, Tian and Fu, Zihao and Cai, Deng and Liu, Lemao and Collier, Nigel and Watanabe, Taro and Su, Yixuan},
  journal={Advances in Neural Information Processing Systems},
  volume={36},
  pages={72888--72903},
  year={2023}
}

@inproceedings{Frustratingly,
  title={A frustratingly simple decoding method for neural text generation},
  author={Yang, Haoran and Cai, Deng and Li, Huayang and Lam, Wai and Shi, Shuming and others},
  booktitle={Proceedings of the 2024 Joint International Conference on Computational Linguistics, Language Resources and Evaluation (LREC-COLING 2024)},
  pages={536--557},
  year={2024}
}

@article{LZPenalty,
  title={LZ Penalty: An information-theoretic repetition penalty for autoregressive language models},
  author={Ginart, Antonio A and Kodali, Naveen and Lee, Jason and Xiong, Caiming and Savarese, Silvio and Emmons, John R},
  journal={arXiv preprint arXiv:2504.20131},
  year={2025}
}

@inproceedings{WangLL0S024,
  title={Mitigating the language mismatch and repetition issues in llm-based machine translation via model editing},
  author={Wang, Weichuan and Li, Zhaoyi and Lian, Defu and Ma, Chen and Song, Linqi and Wei, Ying},
  booktitle={Proceedings of the 2024 Conference on Empirical Methods in Natural Language Processing},
  pages={15681--15700},
  year={2024}
}

@article{EntropyLens,
  title={Entropy-Lens: Uncovering Decision Strategies in LLMs},
  author={Ali, Riccardo and Caso, Francesco and Irwin, Christopher and Li{\`o}, Pietro},
  journal={arXiv preprint arXiv:2502.16570},
  year={2025}
}

@article{mme,
  title={Mme: A comprehensive evaluation benchmark for multimodal large language models},
  author={Fu, Chaoyou and Chen, Peixian and Shen, Yunhang and Qin, Yulei and Zhang, Mengdan and Lin, Xu and Yang, Jinrui and Zheng, Xiawu and Li, Ke and Sun, Xing and others},
  journal={Advances in Neural Information Processing Systems},
  volume={38},
  year={2026}
}

@inproceedings{mathew2021docvqa,
  title={Docvqa: A dataset for vqa on document images},
  author={Mathew, Minesh and Karatzas, Dimosthenis and Jawahar, CV},
  booktitle={2021 IEEE Winter Conference on Applications of Computer Vision (WACV)},
  pages={2199--2208},
  year={2021},
  organization={IEEE}
}

@inproceedings{masry2022chartqa,
  title={Chartqa: A benchmark for question answering about charts with visual and logical reasoning},
  author={Masry, Ahmed and Tan, Jia Qing and Joty, Shafiq and Hoque, Enamul and others},
  booktitle={Findings of the association for computational linguistics: ACL 2022},
  pages={2263--2279},
  year={2022}
}

@article{mmstar,
  title={Are we on the right way for evaluating large vision-language models?},
  author={Chen, Lin and Li, Jinsong and Dong, Xiaoyi and Zhang, Pan and Zang, Yuhang and Chen, Zehui and Duan, Haodong and Wang, Jiaqi and Qiao, Yu and Lin, Dahua and others},
  journal={Advances in Neural Information Processing Systems},
  volume={37},
  pages={27056--27087},
  year={2024}
}

@article{li2023seed,
  title={Seed-bench: Benchmarking multimodal llms with generative comprehension},
  author={Li, Bohao and Wang, Rui and Wang, Guangzhi and Ge, Yuying and Ge, Yixiao and Shan, Ying},
  journal={arXiv preprint arXiv:2307.16125},
  year={2023}
}

@inproceedings{lu2024mathvista,
  title={Mathvista: Evaluating mathematical reasoning of foundation models in visual contexts},
  author={Lu, Pan and Bansal, Hritik and Xia, Tony and Liu, Jiacheng and Li, Chunyuan and Hajishirzi, Hannaneh and Cheng, Hao and Chang, Kai-Wei and Galley, Michel and Gao, Jianfeng},
  booktitle={International Conference on Learning Representations},
  volume={2024},
  pages={23439--23554},
  year={2024}
}

@inproceedings{zhang2024mathverse,
  title={Mathverse: Does your multi-modal llm truly see the diagrams in visual math problems?},
  author={Zhang, Renrui and Jiang, Dongzhi and Zhang, Yichi and Lin, Haokun and Guo, Ziyu and Qiu, Pengshuo and Zhou, Aojun and Lu, Pan and Chang, Kai-Wei and Qiao, Yu and others},
  booktitle={European Conference on Computer Vision},
  pages={169--186},
  year={2024},
  organization={Springer}
}

@inproceedings{liu2024mmbench,
  title={Mmbench: Is your multi-modal model an all-around player?},
  author={Liu, Yuan and Duan, Haodong and Zhang, Yuanhan and Li, Bo and Zhang, Songyang and Zhao, Wangbo and Yuan, Yike and Wang, Jiaqi and He, Conghui and Liu, Ziwei and others},
  booktitle={European conference on computer vision},
  pages={216--233},
  year={2024},
  organization={Springer}
}

@article{Welleck2020Neural,
  title={Neural text generation with unlikelihood training},
  author={Welleck, Sean and Kulikov, Ilia and Roller, Stephen and Dinan, Emily and Cho, Kyunghyun and Weston, Jason},
  journal={arXiv preprint arXiv:1908.04319},
  year={2019}
}

@article{paulus2017deep,
  title={A deep reinforced model for abstractive summarization},
  author={Paulus, Romain and Xiong, Caiming and Socher, Richard},
  journal={arXiv preprint arXiv:1705.04304},
  year={2017}
}

@article{1,
  title={Hallucination begins where saliency drops},
  author={Zhang, Xiaofeng and Zhu, Yuanchao and Gu, Chaochen and Yuan, Xiaosong and Zhao, Qiyan and Cao, Jiawei and Tang, Feilong and Fan, Sinan and Shen, Yaomin and Shen, Chen and others},
  journal={arXiv preprint arXiv:2601.20279},
  year={2026}
}

@article{2,
  title={Coalition-Aware Skill Reliability for Self-Evolving Agents},
  author={Zhao, Qiyan and Zhang, Xiaofeng and Liu, Bo and Chen, Minda and Xiong, Wei and Chen, Jingyang and Ye, Guanting and Yu, Wenhao and Yuan, Xiaosong and Han, Shijie and others},
  journal={arXiv preprint arXiv:2608.22610},
  year={2026}
}

@article{3,
  title={Context-Aware Cluster Decoding: Semantic Anchor-Driven Coherence in dMLLMs},
  author={Zhao, Yikai and Zhao, Qiyan and Zhang, Jiaquan and Zhang, Xiaofeng and Yuan, Xiaosong and Cheng, Pengzhou},
  journal={arXiv preprint arXiv:2608.22367},
  year={2026}
}

@inproceedings{4,
    title = "{T}oken{P}enalty: Alleviating Attention Sinks and Positional Decay in {LVLM}s",
    author = "Zhang, Xiaofeng  and
      Zhu, Yuanchao  and
      Zhao, Qiyan  and
      Yuan, Xiaosong  and
      Cao, Jiawei  and
      Chen, Xuhang",
    booktitle = "Findings of the {A}ssociation for {C}omputational {L}inguistics: {ACL} 2026",
    month = jul,
    year = "2026",
    address = "San Diego, California, United States",
    publisher = "Association for Computational Linguistics",
    pages = "12894--12906",
    ISBN = "979-8-89176-395-1",
}

\makeatletter
\def\@IEEEBIOskipN{1\baselineskip}
\makeatother

\vfill

\end{document}